# A Survey of Multi-Objective Sequential Decision-Making

**Diederik M. Roijers**                                    D.M.ROIJERS@UVA.NL
*Informatics Institute*
*University of Amsterdam*
*Amsterdam, The Netherlands*

**Peter Vamplew**                                    P.VAMPLEW@BALLARAT.EDU.AU
*School of Science,*
*Information Technology and Engineering*
*University of Ballarat*
*Ballarat, Victoria, Australia*

**Shimon Whiteson**                                    S.A.WHITESON@UVA.NL
*Informatics Institute*
*University of Amsterdam*
*Amsterdam, The Netherlands*

**Richard Dazeley**                                    R.DAZELEY@BALLARAT.EDU.AU
*School of Science,*
*Information Technology and Engineering*
*University of Ballarat*
*Ballarat, Victoria, Australia*

## Abstract

Sequential decision-making problems with multiple objectives arise naturally in practice and pose unique challenges for research in decision-theoretic planning and learning, which has largely focused on single-objective settings. This article surveys algorithms designed for sequential decision-making problems with multiple objectives. Though there is a growing body of literature on this subject, little of it makes explicit under what circumstances special methods are needed to solve multi-objective problems. Therefore, we identify three distinct scenarios in which converting such a problem to a single-objective one is impossible, infeasible, or undesirable. Furthermore, we propose a taxonomy that classifies multi-objective methods according to the applicable scenario, the nature of the scalarization function (which projects multi-objective values to scalar ones), and the type of policies considered. We show how these factors determine the nature of an optimal solution, which can be a single policy, a convex hull, or a Pareto front. Using this taxonomy, we survey the literature on multi-objective methods for planning and learning. Finally, we discuss key applications of such methods and outline opportunities for future work.

## 1. Introduction

Sequential decision problems, commonly modeled as *Markov decision processes* (MDPs) (Bellman, 1957a), occur in a range of real-world tasks such as robot control (Kober & Peters, 2012), game playing (Szita, 2012), clinical management of patients (Peek, 1999), military planning (Aberdeen, Thiébaux, & Zhang, 2004), and control of elevators (Crites & Barto, 1996), power systems (Ernst, Glavic, & Wehenkel, 2004), and water supplies (Bhattacharya, Lobbrecht, & Solomantine, 2003). Therefore, the development of algorithms for automatically solving such problems, either by planning given a model of the MDP (e.g., via *dynamic programming* methods, Bellman, 1957b) or by learning through interaction with an unknown MDP (e.g., via *temporal-difference* methods, Sutton & Barto, 1998), is an important challenge in artificial intelligence.

In most research on these topics, the desirability or undesirability of actions and their effects are codified in a single, scalar reward function. Typically, the objective of the autonomous agent interacting with the MDP is then to maximize the expected (possibly discounted) sum of these rewards over time. In many tasks, a scalar reward function is the most natural, e.g., a financial trading agent could be rewarded based on the monetary gain or loss in its holdings over the most recent time period. However, there are also many tasks that are more naturally described in terms of multiple, possibly conflicting objectives, e.g., a traffic control system should minimize latency and maximize throughput; an autonomous vehicle should minimize both travel time and fuel costs. Multi-objective problems have been widely examined in many areas of decision-making (Zeleny & Cochrane, 1982; Vira & Haimes, 1983; Stewart, 1992; Diehl & Haimes, 2004; Roijers, Whiteson, & Oliehoek, 2013) and there is a growing, albeit fragmented, literature addressing multi-objective decision-making in sequential settings.

In this article, we present a survey of the algorithms that have been devised for such settings. We begin in Section 2 by formalizing the problem as a *multi-objective MDP* (MOMDP). Then, in Section 3, we motivate the multi-objective perspective on decision-making. Little of the existing literature on multi-objective algorithms makes explicit why a multi-objective approach is beneficial and, crucially, which cases cannot be trivially reduced to a single-objective problem and solved with standard algorithms. To address this, we describe three motivating scenarios for multi-objective algorithms.

Then, in Section 4, we present a novel taxonomy that organizes multi-objective problems in terms of their underlying assumptions and the nature of the resulting solutions. A key difficulty with the existing literature is that authors have considered many different types of problems, often without making explicit the assumptions involved, how these differ from those of other authors, or the scope of applicability of the resulting methods. Our taxonomy aims to fill this void.

Sections 5 and 6 survey MOMDP planning and learning methods, respectively, organizing them according to the taxonomy and identifying some key differences between the approaches examined in the planning and learning areas. Section 7 surveys applications of these methods, covering both specific applications and more general classes of problems where MOMDP methods can be applied. Section 8 discusses future directions for the field based on gaps in the literature identified in Sections 5 and 6, and Section 9 concludes.

## 2. Background

A finite single-objective *Markov decision process* (MDP) is a tuple $\langle S, A, T, R, \mu, \gamma \rangle$ where:

- $S$ is a finite set of *states*,

- $A$ is a finite set of *actions*,

- $T : S \times A \times S \rightarrow [0, 1]$ is a *transition function* specifying, for each state, action, and next state, the probability of that next state occurring,

- $R : S \times A \times S \rightarrow \Re$ is a *reward function*, specifying, for each state, action, and next state, the expected immediate reward,

- $\mu : S \rightarrow [0, 1]$ is a probability distribution over initial states, and

- $\gamma \in [0, 1)$ is a *discount factor* specifying the relative importance of immediate rewards.

The goal of an agent that acts in this environment is to maximize the expected *return* $R_t$, which is some function of the rewards received from timestep $t$ and onwards. Typically, the return is *additive* (Boutilier, Dean, & Hanks, 1999), i.e., it is a sum of these rewards. In an *infinite horizon* MDP, the return is typically an infinite sum, with each term discounted according to $\gamma$:

$$R_t = \sum_{k=0}^{\infty} \gamma^k r_{t+k+1},$$

where $r_t$ is the reward obtained at time $t$. The parameter $\gamma$ thus quantifies the relative importance of short-term and long-term rewards.

In contrast, in a *finite horizon* MDP, the return is typically an undiscounted finite sum, i.e., after a certain number of timesteps, the process terminates and no more reward can be obtained. While single- and multi-objective methods have been developed for finite horizon, discounted infinite horizon, and average reward settings (Puterman, 1994), for the sake of brevity we formalize only infinite horizon discounted reward MDPs in this article.[1]

An agent's *policy* $\pi$ determines which actions it selects at each timestep. In the broadest sense, a policy can condition on everything that is known to the agent. A *state-indepedent value function* $V^\pi$ specifies the expected return when following $\pi$ from the initial state:

$$V^\pi = E[R_0 \mid \pi]. \tag{1}$$

If the policy is *stationary*, i.e., it conditions only on the current state, then it can be formalized as $\pi : S \times A \rightarrow [0, 1]$: it specifies, for each state and action, the probability of taking that action in that state. We can then specify the *state value function* of a policy $\pi$:

$$V^\pi(s) = E[R_t \mid \pi, s_t = s],$$

for all $t$ when $s_t = s$. The *Bellman equation* restates this expectation recursively for stationary policies:

$$V^\pi(s) = \sum_a \pi(s, a) \sum_{s'} T(s, a, s')[R(s, a, s') + \gamma V^\pi(s')].$$

Note that the Bellman equation, which forms the heart of most standard solution algorithms such as dynamic programming (Bellman, 1957b) and temporal-difference methods (Sutton & Barto, 1998), explicitly relies on the assumption of additive returns. This is important because, as we explain in Section 4.2.2, some multi-objective settings can interfere with this additivity property, making planning and learning methods that rely on the Bellman equation inapplicable.

---

1. For formalizations of the other settings, see for example the overview by Van Otterlo and Wiering (2012).

State value functions induce a partial ordering over policies, i.e., $\pi$ is better than or equal to $\pi'$ if and only if its value is greater for all states:

$$\pi \succeq \pi' \Leftrightarrow \forall s, V^\pi(s) \geq V^{\pi'}(s).$$

A special case of a stationary policy is a *deterministic* stationary policy, in which one action is chosen with probability 1 for every state. A deterministic stationary policy can be seen as a mapping from states to actions: $\pi : S \rightarrow A$. For single-objective MDPs, there is always at least one *optimal* policy $\pi$, i.e., $\forall \pi' : \pi \succeq \pi'$, that is stationary and deterministic.

**Theorem 1.** *For any additive infinite-horizon single-objective MDP, there exists a deterministic stationary optimal policy (see e.g., Howard, 1960; Boutilier et al., 1999).*

If more than one optimal policy exists, they share the same value function, known as the *optimal value function* $V^*(s) = \max_\pi V^\pi(s)$. The *Bellman optimality equation* defines the optimal value function recursively:

$$V^*(s) = \max_a \sum_{s'} T(s, a, s')[R(s, a, s') + \gamma V^*(s')].$$

Note that, because it maximizes over actions, this equation makes use of the fact that there is an optimal deterministic stationary policy. Because an optimal policy maximizes the value for every state, such a policy is optimal regardless of the initial state distribution $\mu$. However, the state-independent value (Equation 1) may very well be different for different initial state distributions. Using $\mu$, the state value function can be translated back into the state-independent value function (Equation 1):

$$V^\pi = \sum_{s \in S} \mu(s) V^\pi(s).$$

A *multi-objective MDP* (MOMDP)[2] is an MDP in which the reward function $\mathbf{R} : S \times A \times S \rightarrow \Re^n$ describes a vector of $n$ rewards, one for each objective, instead of a scalar. Similarly, a value function $\mathbf{V}^\pi$ in an MOMDP specifies the expected cumulative discounted reward vector:

$$\mathbf{V}^\pi = E[\sum_{k=0}^\infty \gamma^k \mathbf{r}_{k+1} \mid \pi], \tag{2}$$

where $\mathbf{r}_t$ is the vector of rewards received at time $t$. The only difference between the single objective value (Equation 1) and the multi-objective value (Equation 2) of a policy is that the return, and the underlying sum of rewards, is now a vector rather than a scalar. For stationary policies, we can also define the multi-objective value of a state:

$$\mathbf{V}^\pi(s) = E[\sum_{k=0}^\infty \gamma^k \mathbf{r}_{t+k+1} \mid \pi, s_t = s]. \tag{3}$$

In a single-objective MDP, state value functions impose only a partial ordering because policies are compared at different states, e.g., it is possible that $V^\pi(s) > V^{\pi'}(s)$ but $V^\pi(s') <$ $V^{\pi'}(s')$. But for a given state, the ordering is complete, i.e., $V^{\pi}(s)$ must be greater than, equal to, or less than $V^{\pi'}(s)$. The same is true of state-independent value functions.

---

2. Multi-objective MDPs should not be confused with *mixed-observability MDPs* (Ong, Png, Hsu, & Lee, 2010), which are also sometimes abbreviated with 'MOMDP'.

In contrast, in an MOMDP, the presence of multiple objectives means that the value function $\mathbf{V}^{\pi}(s)$ for a state $s$ is a vector of expected cumulative rewards instead of a scalar. Such value functions supply only a partial ordering, even for a given state. For example, it is possible that, for some state $s$, $V_i^{\pi}(s) > V_i^{\pi'}(s)$ but $V_j^{\pi}(s) < V_j^{\pi'}(s)$. Similarly, for state-independent value functions, it may be that $V_i^{\pi} > V_i^{\pi'}$ but $V_j^{\pi} < V_j^{\pi'}$. Consequently, unlike in an MDP, we can no longer determine which values are optimal without additional information about how to prioritize the objectives. Such information can be provided in the form of a *scalarization function*, which we discuss in the following sections.

Though not the focus of this article, there are also MOMDP variants in which *constraints* are specified on some objectives (see e.g., Feinberg & Shwartz, 1995; Altman, 1999). The goal of the agent is then to maximize the regular objectives while meeting the constraints on the other objectives. Constrained objectives are fundamentally different from regular objectives because they are explicitly prioritized over the regular objectives, i.e., any policy that fails to meet a constraint is inferior to any policy that meets all constraints, regardless of how well the policies maximize the regular objectives.

## 3. Motivating Scenarios

While the MOMDP setting has received considerable attention, it is not immediately obvious why it is a useful addition to the standard MDP or why specialized algorithms for it are needed. In fact, some researchers argue that modeling problems as explicitly multi-objective is not necessary, and that a scalar reward function is adequate for all sequential decision-making tasks. The most direct formulation of this perspective is Sutton's *reward hypothesis*, which states "that all of what we mean by goals and purposes can be well thought of as maximization of the expected value of the cumulative sum of a received scalar signal (reward)."[3]

This view does not imply that multi-objective problems do not exist. Indeed, that would be a difficult claim, since it is so easy to think of problems that naturally possess multiple objectives. Instead, the implication of the reward hypothesis is that the resulting MOMDPs can always be converted into single-objective MDPs with additive returns. Such a conversion process would involve two steps. The first step is to specify a *scalarization function*.

**Definition 1.** *A* scalarization function $f$, *is a function that projects the multi-objective value $\mathbf{V}^{\pi}$ to a scalar value.*

$$V_{\mathbf{w}}^{\pi}(s) = f(\mathbf{V}^{\pi}(s), \mathbf{w}),$$

*where $\mathbf{w}$ is a weight vector parameterizing $f$.*

For example, $f$ may compute a linear combination of the values, in which case each element of $\mathbf{w}$ quantifies the relative importance of the corresponding objective (this setting is discussed further in Section 4.2.1). The second step is to define a single-objective MDP with additive returns such that, for all $\pi$ and $s$, the expected return equals the scalarized value $V_{\mathbf{w}}^{\pi}(s)$.

---

3. `http://rlai.cs.ualberta.ca/RLAI/rewardhypothesis.html`

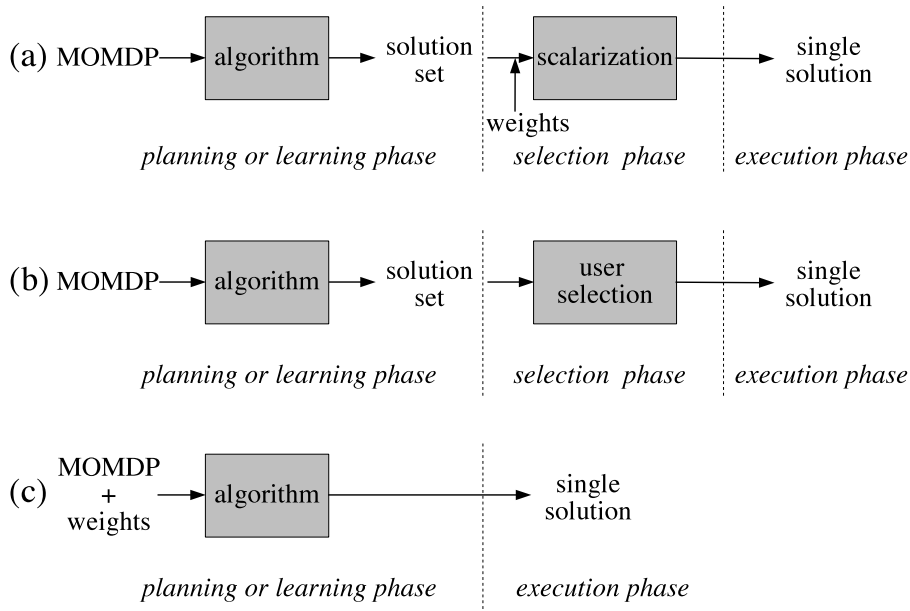

Figure 1: The three motivating scenarios for MOMDPs: (a) the unknown weights scenario, (b) the decision support scenario, (c) the known weights scenario.

Though it rarely, if ever, makes the issue explicit, all research on MOMDPs rests on the premise that there exist tasks for which one or both of these conversion steps is impossible, infeasible, or undesirable. In this section, we discuss three scenarios in which this can occur (see Figure 1).

The first scenario, which we call the *unknown weights scenario* (Figure 1a), occurs when $\mathbf{w}$ is unknown at the moment when planning or learning must occur. Consider for example a public transport system that aims to minimize both latency (i.e., the time that commuters need to reach their destinations) and pollution costs. In addition, assume that the resulting MOMDP can be scalarized by converting each objective into monetary cost: economists can compute the cost of lost productivity due to commuting and pollution incurs a tax that must be paid in pollution credits purchased at a given price. Assume also that those credits are traded on an open market and therefore the price constantly fluctuates. If the transport system is complex, it may be infeasible to compute a new plan every day given the latest prices. In such a scenario, it can be preferable to use a multi-objective planning method that computes a set of policies such that, for any price, one of those policies is optimal (see the *planning or learning phase* in Figure 1a). While doing so is more computationally expensive than computing a single optimal policy for a given price, it needs to be done only once and can be done in advance, when more computational resources are available. Then, when it is time to select a policy, the current weights, i.e., the price of the pollution credits, are used to determine the best policy from the set (the *selection phase*). Finally, the selected policy is employed in the task (the *execution phase*).

In the unknown weights scenario, scalarization is impossible before planning or learning but trivial once a policy actually needs to be used because **w** is known by that time. In contrast, in the second scenario, which we call the *decision support scenario* (Figure 1b), scalarization is infeasible throughout the entire decision-making process because of the difficulty of specifying **w**, or even $f$. For example, economists may not be able to accurately compute the cost of lost productivity due to commuting. The user may also have "fuzzy" preferences that defy meaningful quantification. For example, if the transport system could be made more efficient by building a new train line that obstructs a beautiful view, then a human designer may not be able to quantify the loss of beauty. The difficulty of specifying the exact scalarization is especially apparent when the designer is not a single person but a committee or legislative body whose members have different preferences and agendas. In such a system, the MOMDP method is used to calculate an optimal solution set with respect to the known constraints about $f$ and **w**. As Figure 1b shows, the decision support scenario proceeds similarly to the unknown weights scenario except that, in the selection phase, the user or users select a policy from the set according to their arbitrary preferences, rather than explicit scalarization according to given weights.

In all these cases, one can still argue that scalarization before planning or learning is possible in principle. For example, the loss of beauty can be quantified by measuring the resulting drop in housing prices in neighborhoods that previously enjoyed an unobstructed view. However, the difficulty with scalarization is not only that doing so may be impractical but, more importantly, that it forces the users to express their preferences in a way that may be inconvenient and unnatural. This is because selecting **w** requires weighing hypothetical trade-offs, which can be much harder than choosing from a set of actual alternatives. This is a well understood phenomenon in the field of *decision analysis* (Clemen, 1997), where the standard workflow involves presenting alternatives *before* soliciting preferences. That is why subfields of decision analysis such as *multiple criteria decision-making* and *multi-attribute utility theory* focus on multiple objectives (Dyer, Fishburn, Steuer, Wallenius, & Zionts, 1992). For the same reasons, algorithms for MOMDPs can provide critical decision support. Rather than forcing the users to specify **w** in advance, these algorithms just prune policies that would not be optimal for any **w**. Then, they offer the users a range of alternatives from which they can select according to preferences whose relative importance is not easily quantified.

In the third scenario, which we call the *known weights scenario* (Figure 1c), we assume that **w** is known at the time of planning or learning and thus scalarization is both possible and feasible. However, it may be undesirable because of the difficulty of the second step in the conversion. In particular, if $f$ is nonlinear, then the resulting single-objective MDP may not have additive returns (see Section 4.2.2). As a result, the optimal policy may be non-stationary (see Section 4.3.2) or stochastic (see Section 4.3.3), which cannot occur in single-objective, additive, infinite-horizon MDPs (see Theorem 1). Consequently, the MDP can be difficult to solve, as standard methods are not applicable. Converting the MDP to one with additive returns may not help either as it can cause a blowup in the state space, which also leaves the problem intractable.[4] Therefore, even though scalarization is possible when **w** is known, it may still be preferable to use methods specially designed for MOMDPs rather than to convert the problem to a single-objective MDP. In contrast to the unknown weights and the decision support scenarios, in the known weights scenario, the MOMDP method only produces one policy, which is then executed, i.e., there is no separate selection phase, as shown in Figure 1c.

Note that Figure 1 assumes an *off-line* scenario: planning or learning occurs only once, before execution. However, multi-objective methods can also be employed in *on-line* settings in which planning or learning are interleaved with execution. In the on-line version of the unknown weights scenario, the weights are better characterized as dynamic, rather than unknown. In an on-line scenario, the agent must already have seen weights in all timesteps $t > 1$ since this is a prerequisite for execution in timesteps $1, \ldots, t-1$. However, if the weights change over time, the agent may not yet know the weights that will be used in timestep $t$ when it is in the planning or learning phase of that timestep.

## 4. Problem Taxonomy

So far, we have described the MOMDP formalism and proposed three motivating scenarios for it. In this section, we discuss what constitutes an optimal solution. Unfortunately, there is no simple answer to this question, as it depends on several critical factors. Therefore, we propose a problem taxonomy, shown in Table 1, that categorizes MOMDPs according to these factors and describes the nature of an optimal solution in each category. Our taxonomy is based on what we call the *utility-based approach*, in contrast to many other multi-objective papers that follow an *axiomatic approach* to optimality in MOMDPs.

The utility-based approach rests on the following premise: before the execution phases of the scenarios in Section 3, one policy is selected by collapsing the value vector of a policy to a scalar utility, using the scalarization function. The application of the scalarization function may be implicit and hidden, e.g., it may be embedded in the thought-process of the user, but it nonetheless occurs. The scalarization function is part of the notion of utility, i.e., what the agent should maximize. Therefore, if we find a set with an optimal solution for each possible weight setting of the scalarization function, we have solved the MOMDP. The utility-based approach derives the optimal solution set from the assumptions that are made about the scalarization function, which policies the user allows, and whether we need one or multiple policies.

By contrast, the axiomatic approach begins with the axiom that the optimal solution set is the Pareto front (see Section 4.2.2).[5] This approach is limiting because, as we demonstrate in this section, there are some settings for which other solution concepts are more suitable.

Thus, we take a utility-based approach because it makes it possible to derive the solution concept, rather than just assuming it. When the Pareto front is in fact the correct solution concept, the utility-based approach provides a justification for it. When it is not, it allows for a more appropriate solution concept to be derived instead.

---

4. Since non-additive returns can depend on the agent's entire history, the immediate reward function in the converted MDP may also depend on that history and thus the state representation in the converted MDP must be augmented to include it.

5. For an example of an axiomatic approach to multi-objective reinforcement learning, see the survey by Liu, Xu, and Hu (2013).

|  | single policy (known weights) | | multiple policies (unknown weights or decision support) | |
|---|---|---|---|---|
|  | deterministic | stochastic | deterministic | stochastic |
| linear scalarization | one deterministic stationary policy (1) | | convex coverage set of deterministic stationary policies (2) | |
| monotonically increasing scalarization | one deterministic non-stationary policy (3) | one mixture policy of two or more deterministic stationary policies (4) | Pareto coverage set of deterministic non-stationary policies (5) | convex coverage set of deterministic stationary policies (6) |

Table 1: The MOMDP problem taxonomy showing the critical factors in the problem and the nature of the resulting optimal solution. The columns describe whether the problem necessitates a single policy or multiple ones, and whether those policies must be deterministic (by specification) or are allowed to be stochastic. The rows describe whether the scalarization function is a linear combination of the rewards or, whether this cannot be assumed and the scalarization function is merely a monotonically increasing function of them. The contents of each cell describe what an optimal solution for the given setting looks like.

Our taxonomy categorizes problem classes based on the assumptions about the scalarization function, which policies the user allows, and whether one or multiple policies are required. We show that this leads different solution concepts, underscoring the importance of carefully considering the choice of solution concept based on all the available information.

We discuss the three factors that constitute our taxonomy in the following order. In Section 4.1, we discuss the first factor: whether one or multiple policies are sought, a choice that follows directly from which motivating scenario is applicable. The known weights scenario (Figure 1c) implies a single-policy approach while the unknown weights and decision support scenarios (Figure 1a and 1b) imply a multiple-policy approach. In Section 4.2, we discuss the second factor: whether the scalarization function is a linear combination of the rewards or merely a monotonically increasing function of them. In Section 4.3, we discuss the third factor: whether stochastic or only deterministic policies are permitted.

The goal of the taxonomy is to cover most research on MOMDPs while remaining simple and intuitive. However, due to the diversity of research on MOMDPs, some research does not fit neatly in our taxonomy. We note these discrepancies when discussing such research in Sections 5 and 6.

### 4.1 Single versus Multiple Policies

Following the approach of Vamplew et al. (2011), we first distinguish problems in which only one policy is sought from ones in which multiple policies are sought. Which case holds depends on which of the three motivating scenarios discussed in Section 3 applies.

In the unknown weights and decision support scenarios, the solution to an MOMDP consists of multiple policies. Though these two scenarios are conceptually quite different, from an algorithmic perspective they are identical. The reason is that they are both characterized by a strict separation of the decision-making process into two phases: the planning or learning phase and the execution phase (though in on-line settings, the agent may go back and forth between the two).

In the planning or learning phase, **w** is unavailable. Consequently, the planning or learning algorithm must return not a single policy but a set of policies (and the corresponding multi-objective values). This set should not contain any policies that are suboptimal for all scalarizations, i.e. we are only interested in undominated policies.

**Definition 2.** *For an MOMDP $m$ and a scalarization function $f$, the set of undominated policies, $U(\Pi^m)$, is the subset of all possible policies $\Pi^m$ for $m$ for which there exists a **w** for which the scalarized value is maximal:*

$$U(\Pi^m) = \{\pi : \pi \in \Pi^m \wedge \exists \mathbf{w} \forall (\pi' \in \Pi^m) \ V_{\mathbf{w}}^{\pi} \geq V_{\mathbf{w}}^{\pi'}\}. \tag{4}$$

$U(\Pi^m)$ is sufficient to solve $m$, i.e., for each **w**, it contains a policy with the optimal scalarized value. However, it may contain redundant policies that, while optimal for some weights, are not the *only* optimal policy in the set for **w**. Such policies can be removed while still ensuring the set contains an optimal policy for all **w**. In fact, in order to solve $m$, we need only a subset of the undominated policies such that, for any possible **w**, at least one policy in the set is optimal. This is sometimes called a *coverage set (CS)* (Becker, Zilberstein, Lesser, & Goldman, 2003).

**Definition 3.** *For an MOMDP $m$ and a scalarization function $f$, a set $CS(\Pi^m)$ is a coverage set if it is a subset of $U(\Pi^m)$ and if, for every **w**, it contains a policy with maximal scalarized value, i.e., if:*

$$CS(\Pi^m) \subseteq U(\Pi^m) \wedge (\forall \mathbf{w})(\exists \pi) \left( \pi \in CS(\Pi^m) \wedge \forall (\pi' \in \Pi^m) \ V_{\mathbf{w}}^{\pi} \geq V_{\mathbf{w}}^{\pi'} \right). \tag{5}$$

Note that $U(\Pi^m)$ is automatically a coverage set. However, while $U(\Pi^m)$ is unique, $CS(\Pi^m)$ need not be. When there are multiple policies with the same value, $U(\Pi^m)$ contains all of them, while a coverage set need contain only one. In addition, for a given $CS(\Pi^m)$, there may exist a policy $\pi' \notin CS(\Pi^m)$ for which $\mathbf{V}^{\pi'}$ is different from $\mathbf{V}^{\pi}$ for all $\pi \in CS(\Pi^m)$ but which has the same scalarized value as a $\pi \in CS(\Pi^m)$ for all **w** at which $\pi'$ is optimal. In contrast to single-objective MDPs, in MOMDPs whether or not a policy is in a $CS(\Pi^m)$ can depend on the initial state distribution $\mu$. It is thus important to accurately specify $\mu$ when formulating an MOMDP.

Ideally, an MOMDP algorithm should find the smallest $CS(\Pi^m)$. However, doing so might be harder than just finding one smaller than $U(\Pi^m)$. In Section 4.2, we specialize the coverage set for two classes of scalarization functions.

In the execution phase, a single policy is chosen from the set returned in the planning or learning phase and executed. In the unknown weights scenario, we assume that $\mathbf{w}$ is revealed after planning or learning is complete but before execution begins. Selecting a policy then requires only maximizing over the scalarized value of each policy in the returned set:

$$\pi^* = \operatorname*{argmax}_{\pi \in CS(\Pi^m)} V_{\mathbf{w}}^{\pi}.$$

In the decision support scenario, this set is manually inspected by the user(s), who select a policy for execution informally, making an implicit trade-off between the objectives.

In the known weights scenario, $\mathbf{w}$ is known before planning or learning begins. Therefore, returning multiple policies is unnecessary. However, as mentioned in Section 3 and discussed further in Section 4.2.2, scalarization can yield a single-objective MDP that is difficult to solve.

## 4.2 Linear versus Monotonically Increasing Scalarization Functions

The second critical factor affecting what constitutes an optimal solution to an MOMDP is the nature of the scalarization function. In this section, we discuss two types of scalarization function: those that are linear combinations of the rewards and those that are merely monotonically increasing functions of them.

### 4.2.1 Linear Scalarization Functions

A common assumption about the scalarization function (e.g., Natarajan & Tadepalli, 2005; Barrett & Narayanan, 2008), is that $f$ is linear, i.e., it computes the weighted sum of the values for each objective.

**Definition 4.** *A linear scalarization function computes the inner product of a weight vector* $\mathbf{w}$ *and a value vector* $\mathbf{V}^{\pi}$

$$V_{\mathbf{w}}^{\pi} = \mathbf{w} \cdot \mathbf{V}^{\pi}. \tag{6}$$

*Each element of* $\mathbf{w}$ *specifies how much one unit of value for the corresponding objective contributes to the scalarized value. The elements of the weight vector* $\mathbf{w}$ *are all positive real numbers and constrained to sum to 1.*

Linear scalarization functions are a simple and intuitive way to scalarize. One common situation in which they are applicable is when rewards can be easily translated into monetary value. For example, consider a mining task in which different policies yield different expected quantities of various minerals. If the prices per kilo of those minerals fluctuate daily, then the task can be formulated as an MOMDP, with each objective corresponding to a different mineral. Each element of $\mathbf{V}^{\pi}$ then reflects the expected number of kilos of that mineral that are mined under $\pi$ and the scalarized value $V_{\mathbf{w}}^{\pi}$ corresponds to the monetary value of everything that is mined. $V_{\mathbf{w}}^{\pi}$ can be computed only when $\mathbf{w}$, corresponding to the (normalized) current price per kilo of each mineral, becomes known.

In the single-policy setting, where $\mathbf{w}$ is known, the presence of multiple objectives poses no difficulties given a linear $f$. Instead, $f$ can simply be applied to each reward vector in the MOMDP. Because the inner product computed by $f$ distributes over addition, the result is a single-objective MDP with additive returns. In the infinite horizon setting this leads to:

$$V_{\mathbf{w}}^{\pi} = \mathbf{w} \cdot \mathbf{V}^{\pi} = \mathbf{w} \cdot E[\sum_{k=0}^{\infty} \gamma^k \mathbf{r}_{t+k+1}] = E[\sum_{k=0}^{\infty} \gamma^k (\mathbf{w} \cdot \mathbf{r}_{t+k+1})]. \tag{7}$$

Since this single-objective MDP has additive returns, it can be solved with standard methods, yielding a single policy, as reflected in the box labeled (1) in Table 1. Due to Theorem 1, a determinstic stationary policy suffices. However, a multi-objective approach can still be preferable in this case, e.g., $\mathbf{V}^{\pi}$ may be easier to estimate than $V_{\mathbf{w}}^{\pi}$ in large or continuous MOMDPs where function approximation is required (see Section 6.1).

In the multiple policy setting, however, we do not know $\mathbf{w}$ during planning or learning and therefore want to find a coverage set. If $f$ is linear, then $U(\Pi^m)$, which is automatically a coverage set, consists of the *convex hull*. Substituting Equation 6 in the definition of the undominated set (Definition 2), we obtain the definition of the convex hull:

**Definition 5.** *For an MOMDP $m$, the* convex hull (CH) *is the subset of $\Pi^m$ for which there exists a $\mathbf{w}$ for which the linearly scalarized value is maximal:*

$$CH(\Pi^m) = \{\pi : \pi \in \Pi^m \wedge \exists \mathbf{w} \forall (\pi' \in \Pi^m) \ \mathbf{w} \cdot \mathbf{V}^{\pi} \geq \mathbf{w} \cdot \mathbf{V}^{\pi'}\}. \tag{8}$$

Figure 2a illustrates the concept of a convex hull for stationary deterministic policies. Each point in the plot represents the multi-objective value of a given policy for a two-objective MOMDP. The axes represent the reward dimensions. The convex hull is shown as a set of filled circles, connected by lines that form a convex surface.[6] Given a linear $f$, the scalarized value of each policy is a linear function of the weights. This is illustrated in Figure 2b, where the $x$-axis represents the weight for dimension 0 ($\mathbf{w}[1] = 1 - \mathbf{w}[0]$), and the $y$-axis the scalarized value of the policies. To select a policy, we need only know the values of the convex hull policies, which form the *upper surface* of the scalarized value, as illustrated by the black solid lines, and correspond to the three convex hull policies in Figure 2a. The upper surface forms a piecewise linear and convex function. Such functions are also well-known from the literature on partially-observable Markov decision processes (POMDPs), whose relationship to MOMDPs we discuss in Section 5.2.

Like any $U(\Pi^m)$, $CH(\Pi^m)$ can contain superfluous policies. However, we can also define the *convex coverage set (CCS)* as the specification of the coverage set when $f$ is linear. This is reflected in box (2) in Table 1 (we explain why the policies in this set are deterministic and stationary in Section 4.3.1).

**Definition 6.** *For an MOMDP $m$, a set $CCS(\Pi^m)$ is a* convex coverage set *if it is a subset of $CH(\Pi^m)$ and if, for every $\mathbf{w}$, it contains a policy whose linearly scalarized value is maximal, i.e., if:*

$$CCS(\Pi^m) \subseteq CH(\Pi^m) \wedge (\forall \mathbf{w})(\exists \pi) \left( \pi \in CCS(\Pi^m) \wedge \forall (\pi' \in \Pi^m) \ \mathbf{w} \cdot \mathbf{V}^{\pi} \geq \mathbf{w} \cdot \mathbf{V}^{\pi'} \right). \tag{9}$$

---

6. Note that the term "convex hull" has a slightly different meaning in the multi-objective literature than its standard geometric definition. In geometry, the convex hull of a finite set $S$ of points in Euclidean space is the minimal subset of $S$ so that each of the other points in $S$ can be expressed as a convex combination of the points in the convex hull. In a multi-objective setting, we are only interested in a particular subset of the geometric convex hull; those points of which its convex combinations are strictly bigger (in all dimensions) than any other point in $S$, i.e., all points that are optimal for some weight.

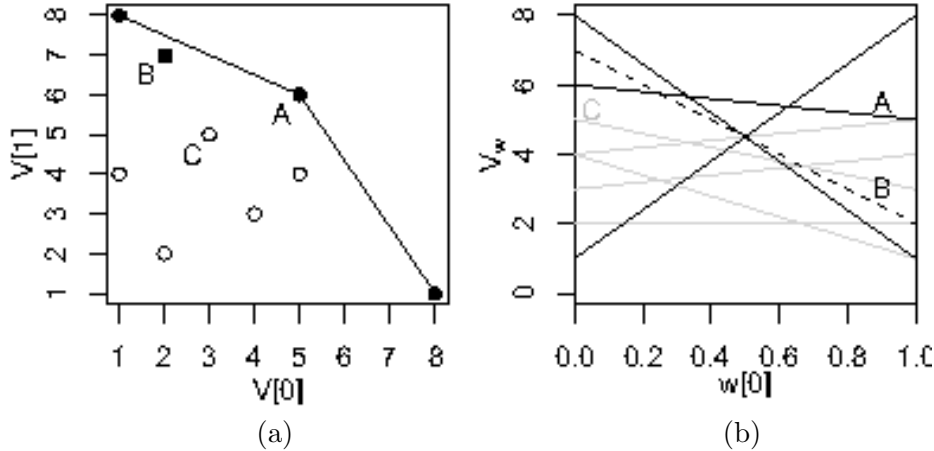

Figure 2: Example of the convex hull and Pareto front. Each point in (a) represents the multi-objective value of a given policy and each line in (b) represents the linearly scalarized value of a policy across values of **w**. The convex hull is shown as black filled circles in (a), and black lines in (b). The Pareto front consists of all filled points (circles and squares) in (a), and both the dashed and solid black lines in (b). The unfilled points in (a) (grey lines in (b)) are dominated.

For deterministic stationary policies, the difference between $CH(\Pi^m)$ and $CCS(\Pi^m)$ may often be small. Therefore, the terms are often used interchangeably. However, in the case of non-stationary or stochastic policies, the difference is quite significant, as the CH can contain infinitely many policies, while it is possible to construct a finite CCS, as we show in Section 4.3.1.

### 4.2.2 Monotonically Increasing Scalarization Functions

While linear scalarization functions are intuitive and simple, they are not always adequate for expressing the user's preferences. For example, suppose in the mining task mentioned above, there are two minerals that can be mined and only three policies are available: $\pi_1$ sends the mining equipment to a location where only the first mineral can be mined, $\pi_2$ to a location where only the second mineral can be mined, and $\pi_3$ to a location where both minerals can be mined. Suppose the owner of the equipment prefers $\pi_3$, e.g., because it at least partially appeases clients with different interests. However, it may be the case that, because the location corresponding to $\pi_3$ has fewer minerals, the convex hull contains only $\pi_1$ and $\pi_2$. Thus, the owner's preference of $\pi_3$ implies that he or she, implicitly or explicitly, employs a nonlinear scalarization function.

Here, we consider the case in which $f$ can be nonlinear, and corresponds to a common notion of the relationship between reward and utility. This class of possibly nonlinear scalarizations are the *strictly monotonically increasing* scalarization functions. These functions adhere to the constraint that if a policy is changed in such a way that its value increases in one or more of the objectives, without decreasing in any other objectives, then the scalarized value also increases.

**Definition 7.** *A scalarization function $f$ is* strictly monotonically increasing *if:*

$$(\forall i, \ V_i^\pi \geq V_i^{\pi'} \wedge \exists i, \ V_i^\pi > V_i^{\pi'}) \Rightarrow (\forall \mathbf{w}, \ V_\mathbf{w}^\pi > V_\mathbf{w}^{\pi'}). \tag{10}$$

Linear scalarization functions (with non-zero positive weights) are included in this class of functions. The condition on the left-hand side of Equation 10 is more commonly known as *Pareto dominance* (Pareto, 1896).

**Definition 8.** *A policy $\pi$* Pareto-dominates *another policy $\pi'$ when its value is at least as high in all objectives and strictly higher in at least one objective:*

$$\mathbf{V}^\pi \succ_P \mathbf{V}^{\pi'} \Leftrightarrow \forall i, V_i^\pi \geq V_i^{\pi'} \wedge \exists i, V_i^\pi > V_i^{\pi'}. \tag{11}$$

Demanding that $f$ is strictly monotonically increasing is quite a minimal constraint, as it requires only that, all other things being equal, getting more reward for a certain objective is always better. In fact, it is difficult to think of any $f$ that violates this constraint without employing a highly unnatural notion of reward.[7]

Three observations are in order about strictly monotonically increasing scalarization functions and the related concept of Pareto dominance. First, unlike in the linear case, we do not necessarily know the exact shape of $f$. Instead, we know only that it belongs to a particular class of functions. The solution concept that follows thus applies to any strictly monotonically increasing $f$. In cases where stronger assumptions about $f$ can be made, more specific solution concepts are possible. However, except for linearity, we are not aware of any such properties of $f$ that have been exploited in solving MOMDPs.

Second, the notions of optimality introduced in Section 4.2.1 are no longer appropriate. The reason is that, even though the vector-valued returns are still additive (Equation 2), the scalarized returns may not be because $f$ may no longer be linear. As an example, consider the well-known *Tchebycheff scalarization function* (Perny & Weng, 2010)[8]:

$$\psi(\mathbf{V}^\pi, \mathbf{p}, \mathbf{w}) = -\max_{i \in 1...n} w_i |p_i - V_i^\pi| - \epsilon \sum_{i \in 1...n} w_i |p_i - V_i^\pi|, \tag{12}$$

where $\mathbf{p}$ is an optimistic *reference point*, $\mathbf{w}$ are weights, and $\epsilon$ is an arbitrarily small positive constant greater than 0. Note that the sum on the righthand side is what makes the function *strictly* monotonically increasing. Now, if $\mathbf{p} = (3, 3)$, $\epsilon = 0.01$, $r_1 = (0, 3)$, $r_2 = (3, 0)$, $\mathbf{w} = (0.5, 0.5)$ and $\gamma = 1$, then $f(\mathbf{V}^\pi, \mathbf{w}) = 0$ but $E[\sum_{k=0}^\infty \gamma^k f(\mathbf{r}_{t+k+1}, \mathbf{w})] = (-1.515) + (-1.515) = -3.03$. This loss of additivity of the scalarized returns when applying a nonlinear $f$ has important consequences for which methods can be applied, as we show in Section 4.3.2.

Third, we can still identify and prune policies that are not optimal for any $\mathbf{w}$ for any strictly monotonically increasing $f$, even though it may be nonlinear. Consider the three labeled policies in Figure 2a (note that Figure 2b does not apply, because the scalarization function is no longer linear). $B$ has a higher value than $A$ in one objective, but a lower value in the other. We therefore cannot tell whether $A$ or $B$ ought to be preferred without knowing $\mathbf{w}$. However, $C$ has a lower value than $A$ in both objectives, and thus $A$ Pareto-dominates $C$: $A \succ_P C$. Because $f$ is strictly monotonically increasing, the scalarized value of $A$ is greater than that of $C$ for all $\mathbf{w}$ and thus we can discard $C$.

---

7. In addition, if $f$ is not strictly monotonically increasing and no other assumptions are made, then no policies can be pruned from the coverage set. Thus, computing the value of every policy in this coverage set, which is required by the selection phase, is likely to be intractable.

8. Our definition differs slightly from that of Perny and Weng (2010): it is multiplied by $-1$ to express maximization instead of minimization, for the sake of consistency with the rest of this article.

For now, we defer a full discussion of what constitutes an optimal solution for an MOMDP with a strictly monotonically increasing scalarization function (i.e., boxes (3)-(6) in Table 1) because this depends, not only on whether the single or multiple policy setting applies, but also on whether only deterministic or also stochastic policies are considered, which is addressed in Section 4.3.

However, we can already observe that, given any strictly monotonically increasing $f$, we can use the *Pareto front* as a set of viable policies. The Pareto front consists of all policies that are not Pareto dominated.

**Definition 9.** *For an MOMDP $m$, the* Pareto front *is the set of all policies that are not Pareto dominated by any other policy in $\Pi^m$:*

$$PF(\Pi^m) = \{\pi : \pi \in \Pi^m \wedge \neg \exists(\pi' \in \Pi^m), \mathbf{V}^{\pi'} \succ_P \mathbf{V}^{\pi}\}. \tag{13}$$

Note that $PF(\Pi^m)$ is not the set of undominated policies $U(\Pi^m)$ for all specific strictly monotonically increasing $f$. We have already seen that for the special case of linear $f$, $U(\Pi^m) = CH(\Pi^m)$, which is a subset of $PF(\Pi^m)$. (For example, in Figure 2, the Pareto front consists of the convex hull plus $B$.) However, for any strictly monotonically increasing $f$, we know that a policy that is not in the $PF(\Pi^m)$ is dominated with respect to $f$, i.e., $\pi \notin PF(\Pi^m) \Rightarrow \pi \notin U(\Pi^m)$. This is because, for strictly monotonically increasing $f$ and $\pi \notin PF(\Pi^m)$, there cannot exist a $\mathbf{w}$ for which $\pi$ is optimal, since by definition there exists a $\pi'$ such that $\mathbf{V}^{\pi'} \succ_P \mathbf{V}^{\pi}$ and, since $f$ is strictly monotonically increasing, this implies that $V^{\pi'}_{\mathbf{w}} > V^{\pi}_{\mathbf{w}}$.

However, if we know only that $f$ is strictly monotonically increasing, we cannot settle for a subset of $PF(\Pi^m)$ either, because there exist strictly monotonically increasing $f$ for which $U(\Pi^m) = PF(\Pi^m)$. Perny and Weng (2010) show that $U(\Pi^m) = PF(\Pi^m)$ for the Tchebycheff function (Equation 12), which is strictly monotonically increasing. Therefore, we cannot discard policies from the $PF(\Pi^m)$ and retain an undominated set $U(\Pi^m)$ for all strictly monotonically increasing $f$.

A *Pareto coverage set* (PCS) of minimal size can be constructed by retaining only one policy of the policies with identical vector values in the $PF(\Pi^m)$. We can formally define the PCS as follows:

**Definition 10.** *For an MOMDP $m$, a set $PCS(\Pi^m)$ is a* Pareto coverage set *if it is a subset of $PF(\Pi^m)$ and if, for every policy $\pi' \in \Pi^m$, it contains a policy that either dominates $\pi'$ or has equal value to $\pi'$, i.e., if:*

$$PCS(\Pi^m) \subseteq PF(\Pi^m) \wedge \forall(\pi' \in \Pi^m)(\exists \pi)\Big(\pi \in PCS(\Pi^m) \wedge (\mathbf{V}^{\pi} \succ_P \mathbf{V}^{\pi'} \vee \mathbf{V}^{\pi} = \mathbf{V}^{\pi'})\Big). \tag{14}$$

Again, for deterministic stationary policies the difference between a $PCS(\Pi^m)$ and $PF(\Pi^m)$ may be minor. Note that $PF(\Pi^m)$ is automatically a $PCS(\Pi^m)$. Most papers in the literature therefore take $PF(\Pi^m)$ as the solution.

We can also slightly relax the constraint on $f$, without having to change which policies are in the $PCS(\Pi^m)$. Specifically, we can define a *monotonically increasing scalarization function* as a function for which the following property holds: $(\forall i, \; V_i^\pi \geq V_i^{\pi'}) \Rightarrow (\forall \mathbf{w}, \; V_\mathbf{w}^\pi \geq V_\mathbf{w}^{\pi'})$. This relaxation influences the set of undominated policies: while policies that are not in the $PF(\Pi^m)$ are always dominated under a strictly monotonically increasing $f$, they need not be under any monotonically increasing $f$. Consider for example $f(\mathbf{V}^\pi, \mathbf{w}) = 0$, which is monotonically increasing but not strictly monotonically increasing. For this function there are no dominated policies, as every policy has the same scalarized value. However, because the scalarized value of a policy $\pi' \notin PF(\Pi^m)$ cannot be *greater* than the scalarized function of a policy $\pi \in PCS(\Pi^m)$, we can use the $PCS(\Pi^m)$ for (non-strict) monotonically increasing $f$. Therefore, in this article, we focus on monotonically increasing $f$, as this is the broader class of functions.

Because the $PF(\Pi^m)$, and even a $PCS(\Pi^m)$, may be prohibitively large and contain many policies whose values differ by negligible amounts, Chatterjee et al. (2006) and Brázdil et al. (2011) introduce a slack parameter $\epsilon$, and use this to define an $\epsilon$-approximate Pareto front, $PF_\epsilon(\Pi^m)$. $PF_\epsilon(\Pi^m)$ contains all values of policies such that for every possible policy $\pi' \in \Pi^m$ there is a policy $\pi \in PF_\epsilon(\Pi^m)$ such that $\forall i \; V_i^\pi(s) + \epsilon \geq V_i^{\pi'}(s)$. By weakening the requirements for domination, this approach yields a smaller set that can be calculated more efficiently.

Another option for finding a smaller set than $PF(\Pi^m)$ is making additional assumptions about the scalarization function. For example, Perny, Weng, Goldsmith, and Hanna (2013) introduce the notion of fairness between objectives, leading to Lorentz optimality. The additional assumption is that if the sum of the values over all objectives stays the same, making the difference between two objectives smaller yields a higher scalarized value. This is of course a strong assumption that does not apply as broadly as Pareto optimality. However, when it does apply, it can help reduce the size of the optimal solution set.

### 4.3 Deterministic versus Stochastic Policies

The third critical factor affecting what constitutes an optimal solution to an MOMDP is whether only deterministic polices are considered or stochastic ones are also allowed. While in most applications there is no reason to exclude stochastic policies a priori, there can be cases when stochastic policies are clearly undesirable or even unethical. For example, if the policy determines the clinical treatment of a patient, e.g., as in work of Lizotte, Bowling, and Murphy (2010) and Shortreed, Laber, Lizotte, Stroup, Pineau, and Murphy (2011), then flipping a coin to determine the course of action may be inappropriate. We denote the set of deterministic policies $\Pi_D^m$ and the set of stationary policies $\Pi_S^m$. Both sets are subsets of all policies: $\Pi_D^m \subseteq \Pi^m \wedge \Pi_S^m \subseteq \Pi^m$. Finally the set of policies that are both deterministic and stationary is the intersection of both these sets, denoted $\Pi_{DS}^m = \Pi_D^m \cap \Pi_S^m$.

In single-objective MDPs, this factor is not critical because, due to Theorem 1, we can restrict our search to deterministic stationary policies, i.e. the optimal attainable value is attainable with a deterministic stationary policy: $\max_{\pi \in \Pi^m} V^\pi = \max_{\pi' \in \Pi_{DS}^m} V^{\pi'}$. However, the situation is more complex in MOMDPs. In this section, we discuss how the focus on stochastic or deterministic policies affects each setting considered in our taxonomy.

### 4.3.1 Deterministic and Stochastic Policies with Linear Scalarization Functions

When $f$ is linear, a result similar to Theorem 1 holds for MOMDPs due to the following corollary:

**Corollary 1.** *For an MOMDP $m$, any $CCS(\Pi_{DS}^m)$ is also a $CCS(\Pi^m)$.*

*Proof.* If $f$ is linear, we can translate the MOMDP to a single-objective MDP, for each possible $\mathbf{w}$. This is done by treating the inner product of the reward vector and $\mathbf{w}$ as the new rewards, and leaving the rest of the problem as is. Since the inner product distributes over addition, the scalarized returns remain additive (Equation 7). Thus, for every $\mathbf{w}$ there exists a translation to a single-objective MDP, for which an optimal deterministic and stationary policy must exist, due to Theorem 1. Hence, for each $\mathbf{w}$ there exists an optimal deterministic stationary policy. Therefore, there exists a $\pi \in CCS(\Pi_{DS}^m)$ that is optimal for that $\mathbf{w}$. Consequently, there cannot exist a $\pi' \in \Pi^m \setminus \Pi_{DS}^m$ such that $\mathbf{w} \cdot \mathbf{V}^{\pi'} > \mathbf{w} \cdot \mathbf{V}^{\pi}$ and thus $CCS(\Pi_{DS}^m)$ is also a $CCS(\Pi^m)$. $\square$

Any $CCS(\Pi_{DS}^m)$ is thus sufficient for solving MOMDPs with linear $f$, even when stochastic and non-stationary policies are allowed. This is reflected in box (2) in Table 1. It also applies to box (1) since the optimal policy in that case is just a member of this $CCS(\Pi_{DS}^m)$, i.e., the one that is best for the given known $\mathbf{w}$.

Unfortunately, no result analogous to Corollary 1 holds for MOMDPs with monotonically increasing $f$. In the rest of this section, we discuss why this is so and the consequences for the nature of an optimal MOMDP solution for boxes (3)-(6) in Table 1.

### 4.3.2 Multiple Deterministic Policies with Monotonically Increasing Scalarization Functions

In the multiple-policy setting when only deterministic policies are allowed and $f$ is nonlinear, non-stationary policies may be better than the best stationary ones.

**Theorem 2.** *In infinite-horizon MOMDPs, deterministic non-stationary policies can Pareto-dominate deterministic stationary policies that are undominated by other deterministic stationary policies (White, 1982).*

To see why, consider the following MOMDP, denoted $m1$, adapted from an example by White (1982). There is only one state and three actions $a_1$, $a_2$, and $a_3$, which yield rewards $(3, 0)$, $(0, 3)$, and $(1, 1)$, respectively. If we allow only deterministic stationary policies, then there are three possible policies $\pi_1, \pi_2, \pi_3 \in \Pi_{DS}^{m1}$, each corresponding to always taking one of the actions, all of which are Pareto optimal. These policies have the following state-independent values (Equation 2): $\mathbf{V}^{\pi_1} = (3/(1-\gamma), 0)$, $\mathbf{V}^{\pi_2} = (0, 3/(1-\gamma))$, and $\mathbf{V}^{\pi_3} = (1/(1-\gamma), 1/(1-\gamma))$. However, if we now consider the set of possibly non-stationary policies $\Pi_D^{m1}$ (including non-stationary ones), we can construct a policy $\pi_{ns} \in \Pi_D^{m1} \setminus \Pi_{DS}^{m1}$ that alternates between $a_1$ and $a_2$, starting with $a_1$, and whose value is $\mathbf{V}^{\pi_{ns}} = (3/(1-\gamma^2), 3\gamma/(1-\gamma^2))$. Consequently, $\pi_{ns} \succ_P \pi_3$ when $\gamma > 0.5$ and thus we cannot restrict our attention to stationary policies.[9] Consequently, in the multiple deterministic policies case with monotonically increasing $f$, we need to find a $PCS(\Pi_D^m)$, which includes non-stationary policies, as shown in box (5) of Table 1.

In addition to having to consider a broader class of policies, another consequence is that defining a policy indirectly via the value function is no longer possible. In standard single-objective methods, the optimal policy can be found by doing *local action selection* with respect to the value function: i.e., for every state, the policy selects the action that maximizes the expected value. However, for local selection to yield a non-stationary policy, the value function must also be non-stationary, i.e., it must condition on the current timestep. While this is standard in the finite-horizon setting, where a different value function is computed for each timestep, it is not possible in the infinite-horizon setting. We discuss how to address this difficulty in Sections 5 and 6.

### 4.3.3 MULTIPLE STOCHASTIC POLICIES WITH MONOTONICALLY INCREASING SCALARIZATION FUNCTIONS

In the multiple policy setting where stochastic non-stationary policies, i.e., the full set $\Pi^m$, are allowed, we again cannot consider only deterministic stationary policies. However, we can employ stochastic stationary policies instead of deterministic non-stationary ones. In particular, we can employ a *mixture policy* (Vamplew, Dazeley, Barker, & Kelarev, 2009) $\pi_m$ that takes a set of $N$ deterministic policies, and selects the $i$-th policy from this set, $\pi_i$ with probability $p_i$, where $\sum_{i=0}^{N} p_i = 1$. This leads to values that are a linear combination of the values of the constituent policies. In our previous example, we can replace $\pi_{ns}$ by a policy $\pi_m$ that chooses $\pi_1$ with probability $p_1$ and $\pi_2$ otherwise, resulting in the following values:

$$\mathbf{V}^{\pi_m} = p_1 \mathbf{V}^{\pi_1} + (1 - p_1)\mathbf{V}^{\pi_2} = \left( \frac{3p_1}{1 - \gamma}, \frac{3(1 - p_1)}{1 - \gamma} \right).$$

Fortunately, it is not necessary to explicitly represent an entire $PCS(\Pi^m)$ explicitly. Instead, it is sufficient to compute a $CCS(\Pi_{DS}^m)$. The necessary stochastic policies to create a $PCS(\Pi^m)$ can then be easily constructed by making mixture policies from those policies on the $CCS(\Pi_{DS}^m)$.

**Corollary 2.** *In an infinite horizon discounted MOMDP, an infinite set of mixture policies $P_M$ can be constructed from policies that are on a $CCS(\Pi_{DS}^m)$, such that this set $P_M$, is a $PCS(\Pi^m)$ (Vamplew et al., 2009).*

*Proof.* We can construct a policy with any value vector on the convex surface, e.g., the black lines in Figure 2a, by mixing policies on a $CCS(\Pi_{DS}^m)$, e.g., the black dots.[10] Therefore, we can always construct a mixture policy that dominates a policy with a value under this surface, e.g., $B$. We can show by contradiction that there cannot be any policy above the convex surface. If there was, it would be optimal for some $\mathbf{w}$ if $f$ was linear. Consequently, by Corollary 1, there would be a deterministic stationary policy with at least equal value. But since the convex surface spans the values on the $CCS(\Pi_{DS}^m)$, this leads to a contradiction. Thus, no policy can Pareto-dominate a mixture policy on the convex surface. □

---

9. White (1982) shows this in an infinite-horizon discounted setting, but the arguments hold also for the finite-horizon and average-reward settings.

10. Note that we should always mix policies that are "adjacent"; the line between any pair of the policies we mix should be on the convex surface. E.g. mixing the policy represented by the leftmost black dot in Figure 2a and the policy represented by the rightmost black dot does not lead to optimal policies, as the line connecting these two points is under the convex surface.

Thanks to Corollary 2, it is sufficient to compute a $CCS(\Pi_{DS}^m)$ to solve an MOMDP, as reflected in box (6) of Table 1. A surprising consequence of this fact, which to our knowledge is not made explicit in the literature, is that Pareto optimality, though the most common solution concept associated with multi-objective problems, is actually only necessary in one specific problem setting:

**Observation 1.** *The multiple policy setting when $f$ is monotonically increasing and only deterministic policies are considered (box (5) of Table 1), requires computing a Pareto coverage set. When either $f$ is linear or stochastic policies are allowed, a $CCS(\Pi_{DS}^m)$ suffices.*

Wakuta (1999) proves the sufficiency of a $CCS(\Pi_{DS}^m)$ for monotonically increasing scalarizations with multiple stochastic policies (box (6) of Table 1) in infinite horizon MOMDPs, but in a different way. Instead of the mixture policies in Corollary 2, he uses stationary randomizations over deterministic stationary policies. Wakuta and Togawa (1998) provide a similar proof for the average reward case.

Note that, while it is common to consider non-stationary or stochastic policies when $f$ is nonlinear, such policies typically condition only on the current state, or the current state and time, not the agent's reward history. However, in this setting, policies that condition on that reward history can dominate those that do not. For example, suppose there are two objectives which can take only positive values and $f$ simply selects the smaller of the two, i.e., $f(\mathbf{V}^\pi, \mathbf{w}) = \min_i V_i^\pi$. Suppose also that, in a given state, two actions are available, which yields rewards of $(4, 4)$ and $(0, 5)$ respectively. Finally, suppose that the agent can arrive at that state with one of two reward histories, whose discounted sums are either $(5, 0)$ or $(3, 3)$. A policy that conditions on these discounted reward histories can outperform policies that do not, i.e., the optimal policy selects the action yielding $(4, 4)$ when the reward history sums to $(3, 3)$ and the action yielding $(0, 5)$ when the reward history sums to $(5, 0)$. So, while for single objective MDPs the Markov property and additive returns are sufficient to restrict our attention to policies that ignore history, in the multi-objective case, the scalarized returns are no longer additive and therefore the optimal policy can depend on the history. Examples of methods that exploit this fact are the 'steering' approach (Mannor & Shimkin, 2001) and the reward-augmented-state thresholded lexicographic ordering method by Geibel (2006), which are discussed in Section 6.1.

### 4.3.4 Single Deterministic and Stochastic Policies with Monotonically Increasing Scalarization Functions

All that remains to address is the single-policy setting with monotonically increasing $f$. The nature of the optimal solution in this case follows directly from the reasoning given for the multiple-policy setting.

If only deterministic policies are considered, then the single policy that is sought may be non-stationary, as reflected in box (3) of Table 1, for the reasons elucidated by White's example. Again, it is hard to define such a non-stationary policy by *local action selection*, due to the risk of circular dependencies in the Q-values.

If stochastic policies are allowed, then the optimal policy may be stochastic, but this can be represented as a mixture policy of two or more deterministic stationary policies, as reflected in box (4) of Table 1, for the same reasons given in Corollary 2. In both cases, policies can potentially benefit from conditioning on the reward history.

## 5. Planning in MOMDPs

In this section, we survey some key approaches to planning in MOMDPs, i.e., computing an optimal policy or the coverage set of undominated policies given a complete model of the MOMDP. Following the taxonomy presented in Section 4, we first consider single-policy methods and then turn to multiple-policy methods for linear and monotonically increasing scalarization functions.

### 5.1 Single-Policy Planning

In the known weights scenario, $\mathbf{w}$ is known before planning begins, and so only a single policy, optimal for $\mathbf{w}$, must be discovered. Since the MOMDP can be transformed to a single-objective MDP when $f$ is linear (see Section 4.2.1), we focus here on single-policy planning for nonlinear $f$.

As discussed in Section 4.2.2, nonlinear $f$ can cause the scalarized return to be non-additive. Consequently, single-objective dynamic programming and linear programming methods, which exploit the assumption of additive returns by employing the Bellman equation, are not applicable. However, different linear programming formulations for single-policy planning in MOMDPs are possible. A key feature of such methods is that they can produce stochastic policies, which, as discussed in Section 4, can be optimal when the scalarization function is nonlinear. While we are not aware of any single-policy planning methods that work for arbitrary nonlinear $f$, methods have been developed for two special cases. In particular, Perny and Weng (2010) propose a linear programming method for MOMDPs scalarized using the Tchebycheff function mentioned in Section 4.2.2. Because the Tchebycheff function always has a $\mathbf{w}$ for which any Pareto-optimal policy is optimal, this approach can find any (single) policy on the Pareto front. In addition, Ogryczak, Perny, and Weng (2011) propose an analogous method for the *ordered weighted regret metric*. This metric calculates the regret of each objective with respect to an estimated ideal reference point, sorts these into descending order, and calculates a weighted sum in which the weights are also in descending order.

Other researchers have proposed single-policy methods for MOMDPs with constraints. Feinberg and Shwartz (1995) consider MOMDPs with one regular objective and $M$ objectives with inequality constraints. They show that if a feasible policy exists for this setting, it can be deterministic and stationary after some finite number of timesteps $N$ and that, prior to timestep $N$, at most $M$ random actions must be performed. They call this a $(M, N)$ policy, show that all Pareto-optimal values can be achieved by $(M, N)$ policies, and propose a linear programming algorithm that finds $\epsilon$-approximate policies for this setting. More general MOMDPs with constraints have also been considered. In particular, Altman (1999) proposes several linear programming approaches for such settings.

Fürnkranz, Hüllermeier, Cheng, and Park (2012) propose a framework for MDPs with qualitative reward signals, which are related to MOMDPs but do not fit neatly in our taxonomy. Qualitative reward signals indicate a preference between policies or actions without directly ascribing a numeric value to them. Since such preferences induce a partial ordering between policies, the policy iteration method the authors propose for this setting may be applicable to MOMDPs with nonlinear $f$, as Pareto dominance also induces partial orderings. However, the authors note that multi-objective tasks generally do have numeric feedback that can be exploited. Thus, they suggest that quantitative MOMDPs can be viewed as a subset of preference-based MDPs, and as such methods designed specifically for MOMDPs may be more efficient than general preference-based methods.

## 5.2 Multiple-Policy Planning with Linear Scalarization Functions

In the multiple-policy setting with linear $f$, we seek a $CCS(\Pi_{DS}^m)$. Note however, the distinction between the convex hull and a convex coverage set is usually not made in the literature.

One might argue that explicitly multi-objective methods are not necessary in this setting, because one could repeatedly run single-objective methods to obtain a $CCS(\Pi_{DS}^m)$. However, since there are infinitely many possible $\mathbf{w}$, it is not obvious that all possible values for $\mathbf{w}$ can be covered. It might be possible to devise a way to run the single-objective methods a finite number times and still guarantee that a $CCS(\Pi_{DS}^m)$ is produced. However, this would be a nontrivial result and the corresponding algorithm would in essence be a multi-objective method that happens to use single-objective methods as subroutines.

One approach that has been attempted to find a minimally sized $CCS(\Pi_D^m)$, i.e., a convex coverage set of deterministic but not necessarily stationary policies, originally proposed by White and Kim (1980), is to translate the MOMDP into a *partially observable Markov decision process* (POMDP) (Sondik, 1971). An intuitive way to think about this translation is to imagine that there is in fact only one true objective but the agent is unaware which of the objectives in the MOMDP it is. This is modeled in the POMDP by defining the state as a tuple $\langle s_1, s_2 \rangle$ where $s_1$ is the state in the MOMDP and $s_2 \in \{1 \ldots n\}$ indicates which is the true objective. The observations thus identify $s_1$ exactly but give no information about $s_2$. Note that this translation from MOMDPs to POMDPs is one-way only. Not every POMDP can be translated to an equivalent MOMDP.

Typically, an agent interacting with a POMDP maintains a *belief*, i.e., a probability distribution over states. In a POMDP derived from an MOMDP, this belief can be decomposed into a belief about $s_1$ and a belief about $s_2$. The former is degenerate because $s_1$ is known. The latter is a vector of size $n$ in which the $i$-th element specifies the probability that the $i$-th objective is the true one. This vector is analogous to $\mathbf{w}$ for a linear $f$. In fact, this is the reason why Figure 2b resembles the piecewise linear value functions often depicted for POMDPs; the only difference is whether the $x$-axis is interpreted as $\mathbf{w}$ or as a belief.

White and Kim (1980) show that, in the finite horizon case, the solution for every belief is exactly the solution for each $\mathbf{w}$, and that the solutions for the resulting POMDP are exactly those for the original MOMDP. The infinite horizon case is more difficult because infinite horizon POMDPs are undecidable (Madani, Hanks, & Condon, 1999). However, for a sufficiently large horizon, the solution to a finite horizon POMDP can be used as an approximate solution to an infinite horizon MOMDP.

To solve the resulting POMDP, White and Kim (1980) propose a combination of Sondik's one-pass algorithm (Smallwood & Sondik, 1973) and policy iteration for POMDPs (Sondik, 1978). However, any POMDP planning method can be used as long as it (1) does not require an initial belief about the POMDP state (which would correspond to initializing not only the MOMDP state but also $\mathbf{w}$) and (2) computes the optimal policy for every possible belief. More recently developed exact methods, e.g., Cassandra, Littman, and Zhang (1997) and Kaelbling, Littman, and Cassandra (1998), meet these conditions and could thus be employed. Approximate point-based POMDP methods (Spaan & Vlassis, 2005; Pineau, Gordon, & Thrun, 2006) do not meet conditions (1) and (2) but could be adapted to compute an approximate convex hull, by choosing a prior distribution for the weights from which they could sample. Online POMDP planning methods (Ross, Pineau, Paquet, & Chaib-draa, 2008) are not applicable because they plan only for a given belief.

Converting to a POMDP thus allows the use of some but not all POMDP methods for solving MOMDPs with linear $f$. However, this approach can be inefficient because it does not exploit the characteristics that distinguish such MOMDPs from general POMDPs, i.e., that part of the state, $s_1$, is known and that no observations give any information about $s_2$. For example, methods that compute policies trees, e.g., (Kaelbling et al., 1998) do not exploit the fact that only deterministic policies that are stationary functions of the state are needed for MOMDPs with linear $f$. Furthermore, as mentioned before, general infinite horizon POMDPs are undecidable, but for MOMDPs it is in fact possible to compute the $CCS(\Pi_{DS}^m)$ exactly.

For these reasons, researchers have also developed specialized planning methods for this setting. Viswanathan, Aggarwal, and Nair (1977) propose a linear programming approach for episodic MOMDPs. Wakuta and Togawa (1998) propose a policy iteration approach that has three phases. The first phase uses policy iteration to narrow down the set of possibly optimal policies. The second phase uses linear programs to check for optimality. Since this does not necessarily give a definitive answer, the third phase uses another linear program to handle any undetermined solutions left after the second phase.

Barrett and Narayanan (2008) propose *convex hull value iteration* (CHVI), which computes the $CH(\overset{\circ}{\Pi}_{DS}^m)$, in every state. CHVI extends conventional value iteration by storing a set of vectors, $\overset{\circ}{Q}(s,a)$ for each state-action pair, representing the convex hull of policies involving that action. These sets of vectors correspond to the Q-values in the single-objective setting; they contain the optimal Q-values for all possible $\mathbf{w}$. When a backup operation is performed, the Q-hulls at the next state $s'$ are propagated back to $s$. For each possible next state $s'$, all possible actions $a'$ are considered (i.e. the union of convex hulls $\bigcup_{a'} \overset{\circ}{Q}(s',a')$ is taken), and weighted by the probability of $s'$ occurring when taking action $a$ in state $s$. This procedure is very similar to the witness algorithm (Kaelbling et al., 1998) for POMDPs.

Lizotte et al. (2010) propose a value-iteration approach for the finite-horizon setting that computes a different value function for each timestep. In addition, it uses a piecewise linear spline representation of the value functions. The authors prove this offers asymptotic time and space complexity improvements over the representation used by CHVI and also enables application of this algorithm to MOMDPs with continuous states. However, the algorithm is only applicable to problems with two objectives. This limitation is addressed in the authors' subsequent work (Lizotte, Bowling, & Murphy, 2012) which extends the algorithm to an arbitrary number of objectives and provides a detailed implementation for the case of three objectives.

## 5.3 Multiple-Policy Planning with Monotonically Increasing Scalarization Functions

In this section, we consider planning in MOMDPs with monotonically increasing $f$. As discussed in Section 4.3, when stochastic policies are allowed, mixture policies of deterministic stationary policies are sufficient. Therefore, we focus on the case when only deterministic policies are allowed and consider methods that compute a $PCS(\Pi_D^m)$, which can include non-stationary policies. The distinction between the $PF(\Pi_D^m)$ and $PCS(\Pi_D^m)$ is usually not made in the literature.

As in the linear case, scalarizing for every $\mathbf{w}$ and obtaining a $PCS(\Pi_D^m)$ by the single-objective methods is problematic. Again there are infinitely many $\mathbf{w}$ to consider but, unlike the linear case, there is the additional difficulty that the scalarized returns may no longer be additive, which can make single-objective methods inapplicable.

Daellenbach and Kluyver (1980) present an algorithm for multi-objective routing tasks (essentially deterministic MOMDPs). Their approach uses dynamic programming in conjunction with an augmented state space to find all non-Pareto-dominated policies iteratively, where the number of iterations equals the maximum number of steps in the route. The algorithm finds undominated sub-policies in parallel. The authors use two alternative explicit scalarization functions, which they call the weighted *minsum* and weighted *minmax* operators. First, the values of all solutions are *translated*: for each objective, the new value becomes the fractional difference between the optimal values for that objective across all solutions. Then, the value for that objective is multiplied by a positive weight. Finally, either the minimum of the sum (*minsum*) or the minimum of the maximal value (*minmax*) of these new weighted fractional differences is chosen as the scalarization. Note that both scalarization functions are monotonically increasing in all objectives, as the optimal value for each objective individually does not depend on the scalarization function.

White (1982) extends this work by proposing a dynamic programming method that approximately solves infinite horizon MOMDPs. It does so by repeatedly backing up according to a multi-objective version of the Bellman equation. Since the policies can be non-stationary, the size of the Pareto front grows rapidly in the number of backups applied. However, White notes that this number "need not be too large before acceptable approximations are reached". Nonetheless, this approach is feasible only for small MOMDPs.

Wiering and De Jong (2007) address this difficulty with a dynamic programming method called CON-MODP for deterministic MOMDPs that computes optimal stationary policies. CON-MODP works by enforcing *consistency* during DP updates: a policy is consistent if it suggests the same action at all timesteps for a given state. If an inconsistent policy is inconsistent only in one state-action pair, CON-MODP makes it consistent by forcing the current action to be taken each time the current state is visited. If the inconsistency runs deeper, the policy is discarded.

By contrast, Gong (1992) proposes a linear programming approach that finds the Pareto-front of stationary policies. However, as the authors note, this approach is also suitable only to small MOMDPs because the number of constraints and decision variables in the linear program increase rapidly as the state space grows.

As mentioned in Section 4.2.2, one way to cope with intractably large Pareto fronts is to compute instead an $\epsilon$-approximate Pareto front, which can be much smaller. Chatterjee et al. (2006) propose a linear programming method that computes the $\epsilon$-approximate front for an infinite horizon MOMDP, while Chatterjee (2007) propose an analogous algorithm for the average reward setting. In both cases, stationary stochastic policies are shown to be sufficient.

Another way to improve scalability in this setting is to give up on planning for the whole state space and instead plan on-line for the agent's current state, using a Monte Carlo tree search approach (Kocsis & Szepesvári, 2006). Such approaches, which have proven very successful, e.g., in the game of Go (Gelly & Silver, 2011), are increasingly popular for single-objective MDPs. Wang and Sebag (2013) propose a Monte Carlo tree search method for deterministic MOMDPs. Single-objective tree search methods typically optimistically explore the tree by selecting actions that maximize the upper confidence bound of their value estimates. The multi-objective variant does the same, but with respect to a scalar multi-objective value function whose definition is based on the *hypervolume indicator* induced by the proposed action together with the set of Pareto optimal policies computed so far. The hypervolume indicator (Zitzler, Thiele, Laumanns, Fonseca, & da Fonseca, 2003) measures the hypervolume that is Pareto-dominated by a set of points. Since the Pareto front maximizes the hypervolume indicator, this optimistic action selection strategy focuses the tree search on the branches most likely to compliment the existing archive.

## 6. Learning in MOMDPs

The methods reviewed in Section 5 assume that a model of the transition and reward dynamics of the MOMDP are known. In cases where such a model is not directly available, *multi-objective reinforcement learning* (MORL) can be used instead.

One way to carry out MORL is to take a *model-based* approach, i.e., use the agent's interaction with the environment to learn a model of the transition and reward function of the MOMDP and then apply multi-objective planning methods such as those described in Section 5. Though such an approach seems well suited to MORL, only a few papers have considered it, (e.g., Lizotte et al., 2010, 2012). We discuss opportunities for future work in model-based MORL in Section 8.1. Instead, most of the work in MORL has focused on *model-free* methods, where a model of the transition and reward function is never explicitly learned.

In this section, we survey some key MORL approaches. While the majority of these methods are for the single-policy setting, multiple-policy methods have also been developed. At first glance, it may seem that multiple-policy methods are unlikely to be effective in the learning setting, since finding more policies would increase sample costs, not just computational costs, and the former is typically a much scarcer resource. However, model-based methods can obviate this issue: once enough samples have been gathered to learn a useful model, finding policies optimal for more weights requires only computation. Model-free methods can also be practical for the multiple-policy setting if they employ *off-policy* learning (Sutton & Barto, 1998; Precup, Sutton, & Dasgupta, 2001), which makes it possible to learn about one policy using data gathered by another. In this way, policies for multiple weight settings can be optimized using the same data.

## 6.1 Single-Policy Learning Methods

In the known weights scenario, a MORL algorithm aims to learn a single policy that is optimal for the given weights. As discussed in Section 5.1, under linear scalarization this is equivalent to learning the optimal policy for a single-objective MDP and so standard *temporal-difference* (TD) methods (Sutton, 1988) such as *Q-learning* (Watkins, 1989) can easily be applied.

However, even though no specialized methods are needed to address this setting, it is nonetheless the most commonly studied setting for MORL. Linear scalarization with uniform weights, i.e., all the elements of $\mathbf{w}$ are equal, forms the basis of the work of Karlsson (1997), Ferreira, Bianchi, and Ribeiro (2012), Aissani, Beldjilali, and Trentesaux (2008) and Shabani (2009) amongst others, while non-uniform weights have been used by authors such as Castelletti et al. (2002), Guo et al. (2009) and Perez et al. (2009). The majority of this work uses TD methods, which work on-line, although Castelletti et al. (2010) extend off-line Fitted Q-Iteration (Ernst, Geurts, & Wehenkel, 2005) to multiple objectives.

In most cases, the only change made to the underlying RL algorithm is that, rather than scalarizing the reward function and then learning a scalar value function in the resulting single-objective MDP, a vector-valued value function is learned in the original MOMDP and then scalarized only when selecting actions. The argument for this approach is that the values of individual objectives may be easier to learn than the scalarized value, particularly when function approximation is employed (Tesauro et al., 2007). For example, each function approximator can ignore any state variables that are irrelevant to its objective, reducing the size of the state space and thereby speeding learning.

As discussed in Section 4.2.2, linear scalarization may not be appropriate for some scenarios. Vamplew, Yearwood, Dazeley, and Berry (2008) demonstrate empirically that this can have practical consequences for MORL. Therefore, MORL methods that can work with nonlinear scalarization functions are of substantial importance. Unfortunately, as illustrated in Section 4.2.2, coping with this setting is especially challenging, since algorithms such as TD methods that are based on the Bellman equation are inherently incompatible with nonlinear scalarization functions due to the non-additive nature of the scalarized returns.

Four main classes of single-policy MORL methods using non-linear scalarization have arisen, which differ in how they deal with this issue. The first class simply applies TD methods without modification. These approaches either resign themselves to being heuristics that are not guaranteed to converge or impose restrictions on the environment to ensure convergence. The second class modifies either the TD algorithm or the state representation such that the issue of non-additive returns is avoided. The third class uses TD methods to learn multiple policies using linear scalarization with different values for $\mathbf{w}$, and then forms a stochastic or non-stationary 'meta-policy' from them that is optimal with respect to a nonlinear scalarization. The fourth class uses policy-search methods, which do not make use of the Bellman equation and hence can be directly applied in combination with nonlinear scalarizations.

The first class includes methods that model the problem as a multi-agent system, with one agent per objective. Each agent learns and recommends actions on the basis of the return for its own objective. A global switch then selects a winning agent, whose recommended action is followed for the current state. Examples include a simple 'winner-takes-all' approach in which the agent whose recommended action has the highest Q-value is selected, or more sophisticated approaches such as W-learning (Humphrys, 1996) where the selected action is the one that will incur the most loss if it is not followed. One key weakness of such approaches was pointed out by Russell and Zimdars (2003): they do not allow for the selection of actions that, while not optimal for any single objective, offer a good compromise between multiple objectives. Another key weakness is that, since the actions selected at different timesteps may be recommended by different agents, the resulting behavior corresponds to a policy that combines elements of those learned by each agent. This combination may not be optimal even for a single objective, i.e., it may be Pareto dominated and perform arbitrarily poorly.

TD has also been used directly with nonlinear scalarization functions that do allow for the consideration of all actions, not just those which are optimal with regards to individual objectives. Scalarization functions based on fuzzy logic have been proposed for problems with discrete actions by Zhao, Chen, and Hu (2010) and for problems with continuous actions by Lin and Chung (1999). A widely cited approach to nonlinear scalarization is that of Gabor, Kalmar, and Szepesvari (1998), which is designed for tasks where constraints must be satisfied for some objectives. A lexicographic ordering of the objectives is defined and a threshold value is specified for all objectives except the last. State-action values for each objective that exceed the corresponding threshold are clamped to that threshold value prior to applying the lexicographic ordering. Thus, this *thresholded lexicographic ordering (TLO)* approach to scalarization maximizes performance on the last objective subject to meeting constraints on the other objectives as specified by the thresholds.

While methods combining TD with nonlinear scalarization may converge to a suitable policy under certain conditions, they can also converge to a suboptimal policy or even fail to converge under other conditions. For example, Issabekov and Vamplew (2012) demonstrate empirically that TLO can fail to converge to a suitable policy for episodic tasks if a constrained objective receives non-zero rewards at any timestep other than the end of the episode. In general, methods based on the combination of TD and nonlinear scalarization must be regarded as heuristic in nature, or applicable only to restricted classes of problems.

The second class avoids the problems caused by non-additive scalarized returns by modifying either the TD algorithm or the state representation. To our knowledge, two approaches proposed by Geibel (2006) to address the limitations of TLO are the only members of this class. Both require that the reward accumulated for each objective over the current episode be stored. In the first algorithm, local decision-making is based on the scalarized value of the sum of the cumulative reward and the current state-action values. This eliminates the problem of non-additive returns, but yields a policy that is non-stationary with respect to the observed state, meaning the algorithm may not converge. The second approach augments the state representation with the cumulative reward. This approach converges to the correct policy but learns slowly, due to the increase in the size of the state space.

The third class uses TD methods only to learn policies based on linear scalarizations. A policy selection mechanism based on a nonlinear scalarization is then used to form a 'meta-policy' from these base policies. The Multiple Directions Reinforcement Learning (MDRL) algorithm of Mannor and Shimkin (2001, 2004) uses such an approach in the context of on-line learning for non-episodic tasks. The user specifies a target region within which the long-term average reward should lie. An initial active policy is chosen arbitrarily and followed until the average reward moves outside of the target region and the agent is in a specified reference state. At this point, the direction from the current average reward vector to the closest point of the target set is calculated, and the policy whose direction best matches this target direction is selected as the active policy. In this way, the average reward is 'steered' towards the user's specified target region. While the underlying base policies utilize linear scalarization, the nature of the policy-selection mechanism means that the overall non-stationary policy formed from these base policies is optimal for the nonlinear scalarization specified by the user's defined target set. Vamplew et al. (2009) suggest a similar approach for episodic tasks, with TD used first to learn policies that are optimal under linear scalarization for a range of different **w**, before a stochastic mixture policy is constructed that is optimal with regards to a nonlinear scalarization.

The fourth class uses *policy-search* algorithms that directly learn a policy without learning a value function. For single-policy MORL, research on policy-search approaches has focused on *policy-gradient* methods (Sutton, McAllester, Singh, & Mansour, 2000; Kohl & Stone, 2004; Kober & Peters, 2011). In such methods, a policy is iteratively adjusted in the direction of the gradient of the value with respect to the parameters (usually probability distributions over actions per state) of a policy. Shelton (2001) proposes an algorithm that first learns the optimal policy for each individual objective. These are used as base policies to form an initial mixture policy that stochastically selects a base policy at the start of each episode. A hill-climbing method based on a weighted convex combination of the normalized objective gradients iteratively improves the mixture policy. This approach does not directly fit our taxonomy because the returns themselves are never scalarized. Instead, the weights are used to find a step direction relative to the current policy parameters. From a practical perspective, its behavior is akin to that of single-policy RL using a nonlinear scalarization function, as it converges to a single Pareto-optimal policy that need not lie on the convex hull. Uchibe and Doya (2009) also propose a policy-gradient method for MORL called Constrained Policy Gradient RL (CPGRL) which uses a gradient projection technique to find policies whose average reward satisfies constraints on one or more of the objectives. Like Shelton's approach, CPGRL learns stochastic policies and works with nonlinear scalarization functions.

## 6.2 Multiple-Policy Learning with Linear Scalarization Functions

In the unknown weights and decision support scenarios, if $f$ is linear, then MORL algorithms aim to learn a CCS of the possible policies. A simple but inefficient approach used by Castelletti et al. (2002) is to run TD multiple times with different values of **w**. In the simplest case, the runs are conducted sequentially to gradually build up an approximate CCS. Natarajan and Tadepalli (2005) showed that this approach can be made more efficient by reusing the policies learned on the earlier runs for the most similar **w**. They show that this improves greatly on sample costs when learning a policy for **w** similar to those already visited in previous runs. However, many samples are typically still required before a good approximate CCS is obtained.

A more sophisticated approach to approximating a convex coverage set is to learn multiple policies in parallel. Several algorithms have been proposed to achieve this within a TD learning framework. The approach of Hiraoka, Yoshida, and Mishima (2009) is similar to the CHVI planning algorithm of Barrett and Narayanan (2008) (see Section 5.2) in that it learns in parallel the optimal value function for all **w**, using a convex hull representation. This approach is prone to infinite growth in the number of vertices in convex hull polygons, and so a threshold margin is applied to the hull representations on each iteration, eliminating points that contribute little to the hull's hypervolume. Hiraoka et al. (2009) present an algorithm to adapt the margins during learning to improve efficiency, but note that many parameters must be tuned for effective performance. Mukai, Kuroe, and Iima (2012) present a similar extension of CHVI to a learning context. They address the problematic growth in the number of values stored by pruning vectors after each Q-value update: a vector is selected at random from the set of vectors stored for the given state-action pair and all others lying within a threshold distance of it are deleted.

The approaches of both Hiraoka et al. (2009) and Mukai et al. (2012) are designed for on-line learning. By contrast, Multi-Objective Fitted Q-Iteration (MOFQI) (Castelletti, Pianosi, & Restelli, 2011, 2012) is an off-line approach to learning multiple policies. MOFQI is a multi-objective extension of the Fitted Q-Iteration (FQI) algorithm (Ernst et al., 2005) which uses a combination of historical data about the single-step transition dynamics of the environment, an initial function approximator, and the Q-learning update rule to construct a dataset that maps state-action pairs to their expected return. This dataset is then used to train an improved function approximator and the process repeats until the values of the function approximator converge. MOFQI provides a computationally efficient extension of FQI to multiple objectives by including **w** in the input to the function approximator and constructing an expanded training data set containing training instances with randomly generated **w**'s. Since the learned function generalizes across weight space in addition to state-action space, it can be used to construct a policy for any **w**.

As discussed in Section 5.3, Lizotte et al. (2010) and Lizotte et al. (2012) describe a value-iteration algorithm to find the convex hull of policies for finite horizon tasks. They note that this method can be applied in a learning context by estimating a model of the state transition probabilities and immediate rewards on the basis of experience of the environment. This approach is demonstrated for the task of analyzing randomized drug trial data by producing these estimates from the historical data gathered during the clinical trials.

## 6.3 Multiple-Policy Learning with Monotonically Increasing Scalarization Functions

If $f$ is nonlinear, then MORL algorithms for the unknown weights and decision support scenarios should aim to learn a PCS. As in the linear scalarization case, the simplest approach is to run single-objective algorithms multiple times with varying **w**. Shelton (2001) demonstrates this approach with a policy-gradient algorithm, while Vamplew et al. (2011) do the same with the TLO method of Gabor et al. (1998). This approach however, requires that $f$ is explicitly known to the learning algorithm, which may be undesirable in the decision support scenario.

To our knowledge, there are currently no methods for learning multiple policies with nonlinear $f$ using a value-function approach. While it might seem possible to adapt convex hull methods such as CHVI by using Pareto-dominance operators in place of convex-hull calculations, doing so is not straightforward. Because the scalarized values of policies in a certain state are non-additive, we cannot restrict ourselves to stationary policies if we want to find all deterministic Pareto-optimal policies (as mentioned in Section 4.3.2). However, for the Bellman equation from CHVI to work, additivity, and the resulting sufficiency of deterministic policies, is required. We discuss options for developing multiple-policy learning methods for nonlinear $f$ in Sections 8.1 and 8.2.

Given the extensive research on both multi-objective evolutionary algorithms (MOEAs) (Coello Coello, Lamont, & Van Veldhuizen, 2002; Tan, Khor, Lee, & Sathikannan, 2003; Drugan & Thierens, 2012) and evolutionary methods for RL (Whiteson, 2012), there is surprisingly little work on evolutionary approaches to MORL. As these methods are population-based, they are well suited to approximating Pareto fronts, and would thus seem a natural fit when $f$ is nonlinear. To our knowledge, Handa (2009b) was the first to apply MOEAs to MORL, by extending Estimation of Distribution (EDA) evolutionary algorithms to handle multiple objectives. EDA-RL (Handa, 2009a) uses Conditional Random Fields (CRF) to represent probabilistic policies. An initial set of policies are used to generate a set of episodes. The best episodes from this set are selected and CRFs that are likely to produce these trajectories are generated. The policies formed from these CRFs then constitute the next generation. Handa (2009b) extends EDA-RL to MOMDPs by using a Pareto-dominance based fitness metric to select the 'best' episodes.

Soh and Demiris (2011) also apply MOEAs to MORL. Policies are represented as Stochastic Finite State Controllers (SFSC) and are optimized using two different MOEAs: NSGA2, a standard evolutionary algorithm, and MCMA, an EDA. The use of SFSCs gives rise to a large search space, necessitating the addition of a local search operator. The local search generates a random $\mathbf{w}$, uses it to scalarize the rewards, and performs gradient-based search on the SFSC. Empirical comparisons on multi-objective variants of three POMDP benchmarks demonstrate that the evolutionary methods are generally superior to the purely local-search approach, and that local search combined with evolution usually outperforms the purely evolutionary methods. This is one of very few papers to directly consider partially observable MOMDPs.

## 7. MOMDP Applications

Multi-objective methods for planning and learning have been employed in a wide range of applications, both in simulation and real-world settings. In this section, we survey these applications. For the sake of brevity, this list is not comprehensive but instead aims to provide an illustrative range of examples. First, we discuss the use of multi-objective methods in specific applications. Second, we discuss research that has identified broader classes of problems in which multi-objective methods can play a useful role.

## 7.1 Specific Applications

An important factor driving interest in multi-objective decision-making is the increasing social and political emphasis on environmental concerns. More and more, decisions must be made that trade off economic, social, and environmental objectives. This is reflected in the fact that a substantial proportion of applications of multi-objective methods have an environmental component.

Perhaps the most extensively researched application is the water reservoir control problem considered by Castelletti et al. (2002), Castelletti, Pianosi, and Soncini-Sessa (2008), Castelletti et al. (2011, 2012) and Castelletti, Pianosi, and Restelli (2013). The general task is to find a control policy for releasing water from a dam while balancing multiple uses of the reservoir, including hydroelectric production and flood mitigation. Management of hydroelectric power production has also been examined by Shabani (2009). Another environmental application is that of forest management to balance the economic benefits of timber harvesting with environmental or aesthetic objectives, which has been demonstrated in simulation by both Gong (1992) and Bone and Dragicevic (2009).

Several researchers have also considered environmentally-motivated applications concerning the management of energy consumption. The SAVES system developed by Kwak et al. (2012) controls various aspects of a commercial building (lighting, heating, air-conditioning, and computer systems) to provide a suitable trade-off between energy consumption and the comfort of the building's occupants. Simulation results indicate that SAVES can reduce energy consumption approximately 30% compared to a manual control system, while maintaining or slightly improving occupant comfort. Both Tesauro et al. (2007) and Liu et al. (2010) consider the problem of controlling a computing server, with the objectives of minimizing both response time to user requests and power consumption. Guo et al. (2009) apply MORL to develop a broker agent in the electricity market. The broker sets caps for a group of agents that sit below it in a hierarchy and manage energy consumption at a device level, and must balance energy cost and system stability.

Shelton (2001) also examines the application of MORL to developing broker agents. However, in this case the agent's task is financial rather than environmental, acting as a market maker that sets buy and sell prices for resources in a market. The aim is to balance the objectives of maximizing profit and minimizing spread (the difference between the buy and sell prices) as that will lead to a larger volume of trades[11].

Computing and communications applications have also been widely considered. Perez et al. (2009) apply MORL to the allocation of resources to jobs in a cloud computing scenario, with the objectives of maximizing system responsiveness, utilization of resources, and fairness amongst different classes of user. Comsa et al. (2012) consider how to maximize system throughput and ensure user equity in the context of a Long Term Evolution mobile communications packet scheduling protocol. Tong and Brown (2002) use constraint-based scalarization to address the tasks of call access control and routing in a broadband multimedia network. Their system aims to maximize profit (a function of throughput) while satisfying constraints on quality of service metrics (capacity constraints and fairness constraints), and uses methods similar to that of Gabor et al. (1998). Zheng, Li, Qiu, and Gong (2012) also use constrained MORL methods to make routing decisions in a cognitive radio network, aiming to minimize average transmission delay while maintaining an acceptably low packet loss rate.

---

11. Shelton's model of the market does not directly model trading volume, and so spread is used as a proxy for volume.

Industrial and mechanical control, an important application for single-objective MDP methods, has also been explored by MOMDP researchers. Aoki, Kimura, and Kobayashi (2004) apply distributed RL to control a sewage flow system, exploiting the system's hierarchical structure to find a solution that minimizes violation of stock levels at each node in the flow system, while smoothing variation in flow at the source. Aissani et al. (2008) apply MORL to maintenance scheduling within a manufacturing plant to minimize the time taken to complete all maintenance tasks and machine downtime. Aissani, Beldjilali, and Trentesaux (2009) build on this work by applying it to a simulation of a real petroleum refinery and demonstrating the ability to adapt to unscheduled corrective maintenance required due to equipment failures. The control of a wet clutch in heavy-duty transmission systems is examined by Van Vaerenbergh et al. (2012). There are twin objectives of minimizing engagement time, while also making the transition smooth.

Robotics are also a popular application for MOMDPs, though most work so far has been in simulation rather than on real robots. Maravall and de Lope (2002) consider the control of a two-limbed brachiating robot, with the objectives of moving in a desired direction while avoiding collisions[12]. Nojima, Kojima, and Kubota (2003) also attempt to balance the objectives of progress to a target and collision avoidance. Their agent makes use of predefined behavioral modules for target tracing, collision avoidance, and wall following, with MORL used to dynamically adjust the weighting of the modules. Meisner (2009) identifies social robots as a promising application of MOMDP methods: their behavior is inherently multi-objective because they must carry out a task without causing anxiety or discomfort for humans.

MORL has also been applied to the control of traffic infrastructure. Yang and Wen (2010) apply it to the control of freeway on-ramps and vehicle management systems, aiming to maximize both the throughput and equity of a freeway system. Multiple agents with shared policies are used, with action selection occurring via negotiation between agents. Similarly, Dusparic and Cahill (2009) apply MORL to control traffic lights at intersections in an urban environment to minimize waiting time of two different classes of vehicles. Yin, Duan, Li, and Zhang (2010) and Houli, Zhiheng, and Yi (2010) also apply MORL to traffic light control. The novelty of their approach lies in considering different objectives based on the current state of the road system; minimizing vehicle stops is prioritized when traffic is free-flowing; minimizing waiting time is emphasized when the system is at medium load; and minimizing queue length at intersections is targeted when the system is congested.

Lizotte et al. (2010, 2012) consider a medical application: prescribing an appropriate drug regime for a patient so as to achieve an acceptable trade-off between the drugs' effectiveness and the severity of its side effects. Their system learns multiple policies based on static data produced during randomized controlled drug trials. The selection of the best treatment for a specific patient is then made by a doctor based on that patient's individual circumstances. This application is an excellent example of a problem where stochastic approaches like mixture policies are inappropriate. A policy that maximizes both symptom relief and also side effects for one patient and then minimizes side effects but also symptom relief for the next patient may appear to give excellent results when averaged across episodes. However, the experience of each individual patient will likely be regarded as undesirable.

---

12. In many robotic applications it may be ideal to avoid collisions completely, but in some environments this may not be possible (e.g., in the presence of moving obstacles whose velocity is both faster than that of the robot, and difficult to predict, such as may be the case for humans or human-controlled vehicles) and so reducing both the likelihood and impact of collisions may be more reasonable than attempting to find a collision-free policy. See for example Holenstein and Badreddin (1991) and Pervez and Ryu (2008).

## 7.2 Applications within Broader Planning and Learning Tasks

In addition to the specific applications discussed above, several authors have identified more general classes of tasks in which multi-objective sequential decision-making can be applied.

### 7.2.1 Probabilistic and Risk-Aware Planning

Cheng, Subrahmanian, and Westerberg (2005) argue that decision-making under uncertainty is inherently multi-objective in nature. Even if there is only a single reward to be considered (such as profit), the environmental uncertainty means that the expected value alone is insufficient to support good decision-making; the decision-maker must also consider the variance of the return. Similarly, Bryce (2008) states that probabilistic planning is inherently multi-objective due to the need to optimize both the cost and the probability of success of the plan. He criticizes approaches that either aggregate these factors or bound one and then optimize the other, arguing in favor of explicitly multi-objective methods. The aptly named "Probabilistic Planning *is* Multi-objective!" paper by Bryce, Cushing, and Kambhampati (2007) demonstrates how this might be achieved, describing a method based on multi-objective dynamic programming over belief states, and a multi-objective extension of the Looping AO* search algorithm to find the set of Pareto-optimal plans. Recent work by Kolobov, Mausam, and Weld (2012) and Teichteil-Königsbuch (2012b) examine the extension of stochastic shortest path (SSP) methods to problems where dead-end states exist. SSP methods assume that at least one policy exists which is guaranteed to reach the goal; in the presence of dead-ends no such policy exists, and so the authors propose algorithms which aim to both maximize the probability of reaching the goal and minimize the cost of the paths found to that goal.

Bryce (2008) notes that a probabilistic plan fails when the environment enters a non-goal absorbing state. Hence, multi-objective probabilistic planning has strong parallels with the research into risk-aware RL carried out by Geibel (2001) and Geibel and Wysotzki (2005), which add a second reward signal indicating the transition of the environment into an error state. Defourny, Ernst, and Wehenkel (2008) also provide useful insights into the incorporation of risk-awareness into MDP methods. They review a range of criteria proposed for constraining risk, and note that many of these are nonlinear and produce non-additive scalarized returns that are incompatible with the local decision-making of methods based on the Bellman equation. They recommend that custom risk-control requirements should be mostly enforced heuristically, by altering policy optimization procedures and checking the compliance of the policies with the initial requirements. Multi-policy MOMDP methods treating risk as an additional objective would satisfy this requirement: having iden-tified the coverage set, any risk-aware metric can then be used to select the best policy. However, some measures of risk may not be expressed directly as discounted cumulative rewards. For example, an agent may wish to minimize the variance in the expected return for a particular reward signal rather than its discounted cumulative value. Methods based on multi-objective probabilistic model checking (Courcoubetis & Yannakakis, 1998; Forejt, Kwiatkowska, Norman, Parker, & Qu, 2011; Forejt, Kwiatkowska, & Parker, 2012; Teichteil-Königsbuch, 2012a), which evaluate whether a system modelled as an MDP satisfies multiple, possibly conflicting, properties, may be suitable for such tasks.

### 7.2.2 Multi-Agent Systems

The use of MDPs within multi-agent systems has been widely explored (Bosoniu, Babuska, & Schutter, 2008), and several authors have proposed approaches that are strongly related to MOMDPs. In a multi-agent system, each agent has its own objective, but for effective overall performance must also consider how its actions will affect the other agents. If the agents are not completely self-interested, then this problem can be framed as an MOMDP by treating the effects on other agents as additional objectives. For example, Mouaddib (2006) uses multi-objective dynamic programming to facilitate cooperation between multiple agents whose underlying goals may be conflicting. For each state-action pair, each agent stores three values: its local utility, the gain other agents receive, and the penalty it inflicts on other agents. The policy for each agent is then established by converting these vector values to regret ratios and applying a leximin ordering to these ratios.

Dusparic and Cahill (2010) compare the application of MORL to multi-agent tasks with other multi-agent methods such as evolutionary and ant-colony algorithms. Dusparic and Cahill (2009) extend the W-Learning algorithm of Humphrys (1996). Each agent learns both local policies (one for each of its own objectives) and remote policies (one for each local policy of each of its neighboring agents). At each timestep, all local policies and all active remote policies of each agent nominate actions, and a winning action is selected by combining action values across all nominating policies. A weighting term is applied to the values of remote policies to determine the level of cooperation each agent offers its neighbor. Experimental results in an urban traffic control simulator show substantial improvement when the level of cooperation is non-zero. This work is similar to that of Schneider, Wong, Moore, and Riedmiller (1999), which addresses the use of multiple agents in a distributed network such as a power distribution grid, where the aim is to maximize a global reward formed from a combination of each agent's local reward. They demonstrate that if each agent focuses only on its own local reward, the policies learned may not maximize the global reward, and that performance is improved by having each agent perform linearly scalarized learning using both its own local reward and the rewards of its neighboring agents.

### 7.2.3 Multi-Objective Optimization using Reinforcement Learning

Reinforcement learning is primarily applied to sequential decision-making tasks in a dynamic environment. However it can also be employed to control search mechanisms for static optimization tasks such as scheduling (Carchrae & Beck, 2005). Multi-objective optimization for static tasks such as design is a well-established field and, while the majority of this work has employed mathematical or evolutionary approaches (Coello Coello et al., 2002), a few authors have explored the application of reinforcement learning in such contexts.

Mariano and Morales (1999, 2000b, 2000a) investigate the use of RL methods (Ant-Q and Q-learning) as a search mechanism for optimization of multi-objective design tasks. The values of the decision variables are considered as the current state, and actions are defined that alter the values of those variables. Multiple agents explore the state space in parallel. Agents are divided into 'families', where each family focuses on a single objective. At the end of an episode, the final states found by each agent are evaluated. Undominated solutions are kept in an archive and the agents that discovered those solutions are rewarded, increasing the likelihood of similar policies being followed in the future. The method is shown to work on a small number of test problems from the evolutionary multi-objective optimization literature. Liao, Wu, and Jiang (2010) apply RL to search for static control settings for a power generation system with the objectives of reducing fuel usage and ensuring voltage stability. They propose an RL algorithm that is formulated specifically for tasks with high-dimensional state spaces, and compare its performance against an evolutionary multi-objective algorithm, finding that the RL method discovers fronts that are both more accurate and better distributed, while also improving the speed of search.

Note that to effectively apply RL to multi-objective optimization, assumptions are usually made about the nature of the environment. For example, Liao et al. (2010) require that each action increases or decreases the value of precisely one state variable. As a result, these methods are likely to have limited applicability to the more general MORL problems described earlier.

# 8. Future Work

In this section, we enumerate some of the possibilities for future research in multi-objective planning and learning.

## 8.1 Model-Based Methods

As mentioned in Section 6, there has been very little work on model-based approaches to MORL. Given the breadth of planning methods for MOMDPs, which could be employed in model-based MORL methods as subroutines, this is surprising. To our knowledge, the only work in this area is that of Lizotte et al. (2010, 2012), where a model of the MOMDP's transition probabilities and reward function is derived from historical data, and then a spline-based multi-objective value iteration approach is applied to that model. In general, learning such models seems only negligibly harder than in the single-objective setting, since estimates of each reward function can just be learned separately. The problem of learning the transition function, generally considered the hard part of model learning, is identical to the single-objective setting. Especially in multiple-policy scenarios, model-based approaches to MORL could greatly reduce sample costs: once the model has been learned, an entire CCS or PCS can be computed off-line, without requiring additional samples.

## 8.2 Learning Multiple Policies with Monotonically Increasing Scalarization Functions using Value Functions

As mentioned in Section 6.3, we are not aware of any methods that use a value function approach to learn multiple policies on the PCS. When stochastic policies are permitted, the problem is easier because we can learn $CCS(\Pi_{DS}^m)$, and use either mixture policies (Vamplew et al., 2009) or stationary randomizations (Wakuta, 1999) of the policies on this CCS (see Section 4.3.3). However, when only deterministic policies are permitted, the problem is more difficult. One option could be to use a finite-horizon approximation to the infinite horizon problem. By planning backwards from the planning horizon, the expected reward of $t$ timesteps to go approximates the infinite-horizon value better and better as $t \to \infty$. As mentioned in Section 5.2, similar approaches have been used in the POMDP setting. Another way to find good approximations to non-stationary policies could be to learn stationary policies (perhaps by extending CON-MDP (Wiering & De Jong, 2007) to the learning setting), and prefix them by $t$ timesteps of non-stationary policy.

## 8.3 Many-Objective Sequential Decision-Making

The majority of the research reviewed in this article, both theoretical and applied, deals with MOMDPs with only a few objectives. This mirrors the state of early evolutionary multi-objective research, which focused almost exclusively on problems with two or at most three objectives. However, over the last decade there has been growing interest in evolutionary methods for so-called *many-objective* problems, which have at least four and sometimes more than fifty objectives (Ishibuchi, Tsukamoto, & Nojima, 2008). This research has shown that many algorithms that perform well for a few objectives scale poorly in the number of objectives, necessitating special algorithms for the many-objective setting.

While many-objective MDPs have received little consideration so far, there are numerous real-world control problems that can be naturally modeled in this way. For example, Fleming et al. (2005) point out that many-objective control problems commonly arise in engineering, and give the example of a jet engine control system with eight objectives. As with the many-objective problems considered in evolutionary computation, it seems likely that at least some of the methods explored so far will scale poorly in the number of objectives. For example, the multi-policy MOMDP planning algorithm described by Lizotte et al. (2010) is limited to problems with two objectives.

A key challenge posed by many-objective problems is that the number of undominated solutions typically grows exponentially in the number of objectives. This is particularly problematic for multiple-policy MOMDP methods. Fleming et al. (2005) note that one of the most effective approaches used in many-objective evolutionary computation is to incorporate user preferences to restrict the search space to a small region of interest. In particular, they recommend interactive preference articulation in which the user interactively steers the system towards a desirable solution during optimization. While Vamplew et al. (2011) raise the possibility of incorporating such an approach into MORL, we are not aware of any research that has actually done so.

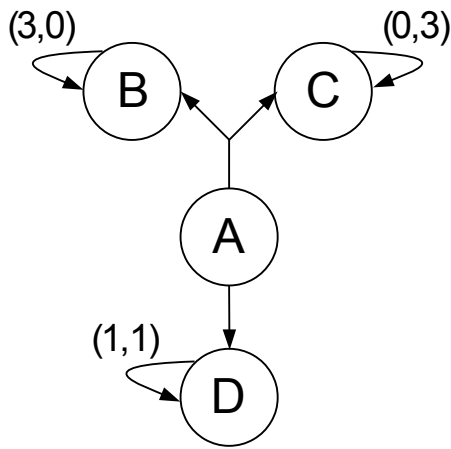

Figure 3: An MOMDP with two objectives and four states.

## 8.4 Expectation of Scalarized Return

In Section 3, we defined the scalarized value $V_\mathbf{w}^\pi(s)$ to be the result of applying the scalarization function $f$ to the multi-objective value $\mathbf{V}^\pi(s)$ according to $\mathbf{w}$, i.e., $V_\mathbf{w}^\pi(s) = f(\mathbf{V}^\pi(s), \mathbf{w})$. Since $\mathbf{V}^\pi(s)$ is itself an expectation, this means that the scalarization function is applied *after* the expectation is computed, i.e.,

$$V_\mathbf{w}^\pi(s) = f(\mathbf{V}^\pi(s), \mathbf{w}) = f(E[\sum_{k=0}^\infty \gamma^k \mathbf{r}_k \mid \pi, s_0 = s], \mathbf{w}).$$

This formulation, which we refer to as the *scalarization of the expected return (SER)* is standard in the literature. However, it is not the only option. It is also possible to define $V_\mathbf{w}^\pi(s)$ as the *expectation of the scalarized return (ESR)*:

$$V_\mathbf{w}^\pi(s) = E[f(\sum_{k=0}^\infty \gamma^k \mathbf{r}_k, \mathbf{w}) \mid \pi, s_0 = s]$$

Which definition is used can critically affect which policies are preferred. For example, consider the following MOMDP, illustrated in Figure 3. There are four states ($A$, $B$, $C$, and $D$) and two objectives. The agent starts in state $A$ and has two possible actions: $a_1$ transits to state $B$ or $C$, each with probability of 0.5, and $a_2$ transits to state $D$ with probability 1. Both actions lead to a $(0,0)$ reward. In states $B$, $C$ and $D$ there is only one action, which leads to a deterministic reward of $(3,0)$ for $B$, $(0,3)$ for $C$, and $(1,1)$ for $D$.

The scalarization function just multiplies the two objectives together. Thus, under SER,

$$V_\mathbf{w}^\pi(s) = V_1^\pi(s)V_2^\pi(s),$$

and under ESR,

$$V_\mathbf{w}^\pi(s) = E\big[\big(\sum_{k=0}^\infty \gamma^k r_k^1\big)\big(\sum_{k=0}^\infty \gamma^k r_k^2\big) \mid \pi, s_0 = s\big],$$

where $r_k^i$ is the reward for the $i$-th objective on timestep $k$ ($\mathbf{w}$ is not needed in this example since $f$ involves no constants). If $\pi_1(A) = a_1$ and $\pi_2(A) = a_2$, then the multi-objective values are $\mathbf{V}^{\pi_1}(A) = (1.5\gamma/(1-\gamma), 1.5\gamma/(1-\gamma))$ and $\mathbf{V}^{\pi_2}(A) = (\gamma/(1-\gamma), \gamma/(1-\gamma))$.

Under SER, this leads to scalarized values of $V^{\pi_1}(A) = (1.5\gamma/(1-\gamma))^2$ and $V^{\pi_2}(A) = (\gamma/(1-\gamma))^2$ and consequently $\pi_1$ is preferred. Under ESR, however, we have $V^{\pi_1}(A) = 0$ and $V^{\pi_2}(A) = (\gamma/(1-\gamma))^2$ and thus $\pi_2$ is preferred.

Intuitively, the SER formulation is appropriate when the policy will be used many times and *return accumulates across episodes*, e.g., because the same user is using the policy each time. Then, scalarizing the expected reward makes sense and $\pi_1$ is preferable because in expectation it will accumulate more return in both objectives. However, if the policy will only be used a few times or the return does not accumulate across episodes, e.g., because each episode is conducted for a different user, then the ESR formulation is more appropriate. In this case, the expected return before scalarization is not of interest and $\pi_2$ is preferable because $\pi_1$ will always yield zero scalarized return on any given episode.

To our knowledge, there is no literature on MOMDPs that employs the ESR formulation, even though there are many real-world scenarios in which it seems more appropriate. For example, in the medical application of Lizotte et al. (2010) mentioned in Section 7, each patient gets only one episode to treat his or her illness, and is thus clearly interested in maximizing ESR, not SER. Thus, we believe that developing methods for MOMDPs under the ESR formulation is a critical direction for future research.

## 9. Conclusions

This article presented a survey of algorithms designed for sequential decision-making problems with multiple objectives.

In order to make explicit under what circumstances special methods are needed to solve multi-objective problems, we identified three distinct scenarios in which converting such a problem to a single-objective one is impossible, infeasible, or undesirable. As well as providing motivation for the need for multi-objective methods, these scenarios also represent the three main ways these methods are applied in practice.

We proposed a taxonomy that classifies multi-objective methods according to the applicable scenario, the scalarization function (which projects multi-objective values to scalar ones), and the type of policies considered. We showed how these factors determine the nature of an optimal solution, which can be a single policy, or a coverage set (convex or Pareto). Our taxonomy is based on a *utility-based approach*, which sees the scalarization function as part of the utility, and thus part of the problem definition. This contrasts with the so-called *axiomatic approach*, which usually assumes the Pareto front is the appropriate solution. We showed that the utility-based approach can be used to justify the choice for a solution set. Following this line of thought, we observed (Observation 1) that computing the Pareto front is often not necessary, and that in many cases a convex coverage set of deterministic stationary policies is sufficient.

Using our taxonomy, we surveyed the literature on multi-objective methods for planning and learning. An interesting observation is that most of the learning methods use a model-free rather than a model based approach, identifying the latter as an understudied class of methods. Another part of the taxonomy which has not yet been widely studied is learning in the case of monotonically increasing scalarization functions.

We discussed key applications of MOMDP methods as motivation for the importance of such methods. Applications were identified in a diverse range of fields including environmental management, financial markets, information and communications technology, and control of industrial processes, robotic systems and traffic infrastructure. In addition connections were identified between multi-objective sequential decision-making and other broad areas of research such as probabilistic planning and model-checking, multi-agent systems and more general multi-objective optimization.

Finally, we outlined several opportunities for future work, which include understudied areas (model-based methods, learning in monotonically increasing scalarization settings, and many-objective sequential decision-making), and a reformulation of the objective for MOMDPs – the Expectation of Scalarized Return – which is particularly important to optimize when a policy can be executed only once.

## Acknowledgments

We would like to thank Matthijs Spaan, Frans Oliehoek, Matthijs Snel, Marie D. Manner and Samy Sá, as well as the anonymous reviewers, for their valuable feedback. This work is supported by the Netherlands Organisation for Scientific Research (NWO): Decision-Theoretic Control for Network Capacity Allocation Problems (#612.001.109) project.